\documentclass{article} 
\usepackage{iclr2025_conference,times}

\usepackage{amsmath,amsfonts,bm}

\def\eqref#1{equation~\ref{#1}}

\def\1{\bm{1}}

\DeclareMathAlphabet{\mathsfit}{\encodingdefault}{\sfdefault}{m}{sl}
\SetMathAlphabet{\mathsfit}{bold}{\encodingdefault}{\sfdefault}{bx}{n}

\usepackage{hyperref}
\usepackage{url}

\usepackage{times}
\usepackage{latexsym}
\usepackage{amsmath}
\usepackage{booktabs}
\usepackage{xcolor}
\usepackage{tcolorbox}
\usepackage{array}
\usepackage{booktabs}
\usepackage{csquotes}
\usepackage{graphicx}
\usepackage{ulem} 
\usepackage{wrapfig}
\usepackage{array}
\usepackage[capitalize,noabbrev]{cleveref}
\usepackage{xspace}
\usepackage{lipsum}
\usepackage{multicol}
\usepackage{amssymb}
\usepackage{textcomp}
\usepackage{colortbl}
    
\title{Understanding the Interplay between Parametric and Contextual Knowledge for Large Language Models}

\author{Sitao Cheng\textsuperscript{$\heartsuit$}\thanks{Work done during a visit to UCSB.},~~Liangming Pan\textsuperscript{$\spadesuit$}\thanks{Corresponding Author},~~Xunjian Yin\textsuperscript{$\heartsuit$}\footnotemark[1],~~Xinyi Wang\textsuperscript{$\heartsuit$},~~William Yang Wang\textsuperscript{$\heartsuit$} \\
\textsuperscript{$\heartsuit$}University of California, Santa Barbara ~~~ \textsuperscript{$\spadesuit$}University of Arizona \\
\texttt{\{sitaocheng,yinxunjian,xinyi\_wang,william\}@ucsb.edu}\\
\texttt{liangmingpan@arizona.edu}\\
}

\newcommand{\ck}{\texttt{CK}\xspace} 
\newcommand{\pk}{\texttt{PK}\xspace}

\iclrfinalcopy 
\begin{document}

\maketitle

\begin{abstract}
Large language models (LLMs) encode vast amounts of knowledge during pre-training (parametric knowledge, or \pk) and can further be enhanced by incorporating contextual knowledge (\ck).
Can LLMs effectively integrate their internal \pk with external \ck to solve complex problems?
In this paper, we investigate the dynamic interaction between \pk and \ck, categorizing their relationships into four types: \textit{Supportive, Complementary, Conflicting}, and \textit{Irrelevant}. To support this investigation, we introduce \textsc{EchoQA}, a benchmark spanning scientific, factual, and commonsense knowledge. Our results show that LLMs tend to suppress their \pk when contextual information is available, even when it is complementary or irrelevant. While tailored instructions can encourage LLMs to rely more on their \pk, they still struggle to fully leverage it. These findings reveal a key vulnerability in LLMs, raising concerns about their reliability in knowledge-intensive tasks. {Resources are available at \href{https://github.com/sitaocheng/Knowledge_Interplay}{https://github.com/sitaocheng/Knowledge\_Interplay}}.

\end{abstract}

\section{Introduction}

\begin{displayquote}
\vspace{-0.1cm}
``\textit{ If a man keeps cherishing his old knowledge, so as continually to be\\ acquiring new.} '' \hfill ---~\textit{Confucius}
\vspace{-0.2cm}
\end{displayquote}

Large language models (LLMs)
\citep{NEURIPS2020_1457c0d6,ouyang2022training,touvron2023llama,zeng2023glmb,openai2023gpt4} capture a substantial amount of knowledge through pre-training on large-scale corpus, referred to as \textit{parametric knowledge} (\pk). When applied to problem-solving, LLMs often rely on additional information provided as context, known as \textit{contextual knowledge} (\ck)\footnote{For simplicity, \pk and \ck denote parametric and contextual knowledge throughout this paper.}. 
A critical area that remains underexplored is how effectively LLMs can integrate \pk and \ck to solve complex problems. This ability is crucial for human-like intelligence, where we effortlessly draw upon stored knowledge to adapt to new environments, reflecting our capacity to \textit{perceive, recall, and integrate} information when encountering new stimuli \citep{gibson1988exploratory}. Effective collaboration between \pk and \ck is also vital for real-world applications, such as retrieval-augmented generation (RAG)~\citep{schick2023toolformer,trivedi-etal-2023-interleaving,an2024threadlogicbaseddataorganization,gutiérrez2024hipporag}. 



\begin{figure*}[t]
\setlength{\abovecaptionskip}{0pt}
\setlength{\belowcaptionskip}{-13pt}
  \includegraphics[width=\linewidth]{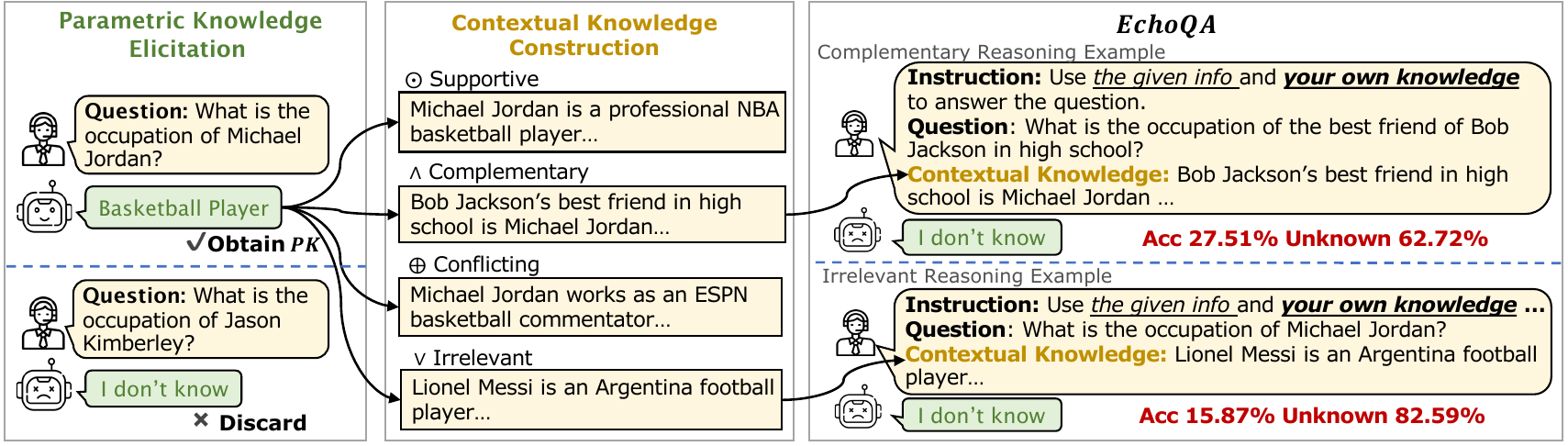}
  \caption {Our benchmark \textsc{EchoQA}, accessing LLMs ability to echo their parametric knowledge (\pk) when contextual knowledge (\ck) is present. We firstly question LLMs to obtain \pk and discard knowledge they cannot answer. Then, we construct \ck~by various \textit{reasoning types} (Table \ref{tab:type_and_metrics}). Next, we question LLMs given \ck.
  Exemplar result is by Llama 3.1-70B on ALCUNA \citep{yin-etal-2023-alcuna}.}

  \label{fig:main_example}
\end{figure*}

However, \textit{we find that for LLMs, an effective interaction between \pk and \ck can be challenging.} For example, we observe a counter-intuitive phenomenon that when supplementing an LLM with more task-relevant knowledge in the context (\ck), sometimes the performance degrades because the LLM fails to effectively utilize the parametric knowledge (\pk). Examples are shown in \cref{fig:main_example}. This catastrophic test time forgetting greatly limits LLMs capability to effectively solve user queries when relevant context is provided. In this paper, our objective is to thoroughly investigate \textit{the intricate dynamics between \pk~and \ck}, leading to a better understanding of how well LLMs integrate and prioritize these sources of knowledge. Previous works separately study when \ck is conflicting or irrelevant to \pk, raising concerns about the vulnerability \citep{wang2023resolving,xie2024knowledgeconflict,wu2024easily}. However, they fail to answer more fine-grained questions such as: To what extent are LLMs aware of \pk, in the presence of \ck? How well can LLMs leverage \pk~given \ck? What factors can affect such ability?

To perform a systematic study, we formulate \textit{reasoning types} based on various relationships between \ck~and \pk\textit{—Supportive, Complementary, Conflicting} and \textit{Irrelevant}, illustrated by examples in Figure \ref{fig:main_example} and Table \ref{tab:type_and_metrics}.
Due to the lack of {appropriate} datasets to examine this ability, we introduce a new benchmark \textbf{\textsc{EchoQA}} (Section \ref{sec:dataset}) spanning diverse knowledge-intensive reasoning tasks, \textit{i.e.,} scientific, factual and commonsense, across various LLMs. We extract \pk~by ensuring LLMs can reach 100\% performance for the knowledge.
Next, we construct \ck~from \pk~to fit our reasoning types, as in the middle of Figure \ref{fig:main_example}.
We hope that \textsc{EchoQA} will serve as a valuable resource to steer future explorations on LLMs leverage and integration of \pk~with \ck.

To investigate to what the extent LLMs can leverage \pk, we design a series of reasoning instructions,
which progressively enforce the use of \pk, guiding LLMs preference to varying degrees, as detailed in Section \ref{sec:reasoning_instructions}.
Furthermore, we investigate other factors
affecting LLMs abilities to leverage their internal knowledge, \textit{e.g.,} popularity of entities and methods of new knowledge fabrication, shedding lights on future improvements (Section \ref{sec:discussion}).
Our investigation draws the following findings:

\noindent  $\bullet$ \textit{The suppression of \pk~in the presence of \ck~is universal}, regardless of the model, knowledge type or reasoning type (Complementary, Conflicting or Irrelevant). LLMs
\textit{disregard} their own knowledge, \textit{e.g.,} more than 60\% of cases in scientific knowledge for all tested models, relying solely on the context for reasoning. This highlights the vulnerability of LLMs in leveraging \pk.

\noindent  $\bullet$ LLMs are more likely to recall their knowledge for some knowledge and reasoning types, \textit{e.g.,} the commonsense knowledge. We find evidence indicating that the reason behind is likely the imbalance of knowledge in training corpus. 

\noindent  $\bullet$ \textit{Explicit instruction can help LLMs remember more \pk,
but still way off from fully leveraging \pk.} This implies more sophisticated prompt or framework design has the potential to solve this problem to a larger extent. We show some possible prompt templates in Appendix \ref{tab:detailed_prompt_reasoning_instructions}.

\section{Related Work}

\textbf{Parametric Knowledge in LLMs} is encoded in models parameters through vast amounts of text data during pre-training. Previous research extensively explores how LLMs leverage \pk~for reasoning \citep{li2024understandingpatchingcompositionalreasoning,yang2024largelanguagemodelslatently,wang2024grokkedtransformersimplicitreasoners}. Some reveal the challenges in leveraging \pk \citep{wang2023survey,allen2023physics}. These studies typically focus on models ability to utilize \pk~to perform various tasks, \textit{e.g.,} multi-hop reasoning, without considering the influence of contextual knowledge (\ck). While valuable, these analyses do not address how \pk~is dynamically applied in real-world contexts where \ck also plays an important role. Moreover, \pk~embeded in language models can be outdated or incorrect, prone to hallucination \citep{elazar-etal-2021-measuring,lazaridou2021mind,zhong-etal-2023-mquake}. Though some work propose to edit \pk~\citep{dai-etal-2022-knowledge,meng2022locating,meng2023massediting}, additional effort would be needed for the loss of other abilities.

\noindent\textbf{Contextual Knowledge for LLMs}
refers to the information (or tools) augmented in LLMs context window to enhance generation (RAG), which hopefully supplements LLMs internal \pk~to improve faithfulness \citep{REALM,Qin2023ToolLW,mallen-etal-2023-trust}. Previous work primarily focuses on optimizing the retrieval of relevant documents for reasoning \citep{press-etal-2023-measuring,asai2023selfrag,zhuang2024efficientragefficientretrievermultihop,gutiérrez2024hipporag}. However, as documents contain diverse information, LLMs may be confused when reasoning, even with optimal retrieval, \textit{e.g.,} when the given knowledge contradicts their knowledge \citep{xu2024knowledgeconflictsllmssurvey}. We find that even complementary \ck~can hinder LLMs ability to fully leverage their knowledge. In this paper, we formulate the relationship between \pk~and \ck~to investigate the influence of different kinds of \ck~on LLMs utilization of \pk.

\noindent\textbf{Interplay between Parametric and Contextual Knowledge} mainly includes behavioral analyses or fusion of both knowledge.
Some assess models handling of new knowledge \citep{yin-etal-2023-alcuna}, focusing on knowledge construction. Other works elicit \pk~from LLMs and design frameworks to introduce conflict \citep{wang2023resolving,xie2024knowledgeconflict} or irrelevant \ck \citep{wu2024easily}.
However, the effectiveness of LLMs leveraging of \pk under various \ck is understudied. Besides, the focus on single type of knowledge may overlook the complex relation between the two knowledge sources
reflected real-world scenarios, \textit{e.g.,} LLM-based search engines or chatbots like ChatGPT Plugins and New Bing \citep{nakano2021webgpt,deng2023mind2web} or human reasoning process, \textit{e.g.,} when exploring new environments, human intelligence can naturally integrate knowledge in brain with outside nature. While some studies explore the fusion of \ck~and \pk~for improved sufficiency \citep{jiang-etal-2023-active,zhang2024evaluating}, the intricate relationship between them is understudied. We aim to comprehensively examine LLMs ability to manage and integrate these two knowledge sources.

\section{Experimental Setup}
\label{sec:experimental_setup}




\begin{table*}[t]
\centering
\setlength{\abovecaptionskip}{0pt}
\setlength{\belowcaptionskip}{0pt}
\caption{Definition, example and metric of our reasoning types over various relationships between \ck~and \pk. Note that we ensure \pk~is known by LLMs and not shown in context. $U_R, M_R, A_{ck}, A_{pk}$ denotes unknown ratio, memorization ratio, answers faithful to \ck and \pk, respectively.}
\vspace{1pt} 
\begin{tabular}{p{2.4cm} p{8.7cm} p{1.5cm}} 
\toprule
\textbf{Reasoning Type} & \textbf{Example}  & \textbf{Metric} \\
\midrule
\textit{Supportive} \newline \(\ck \odot \pk~\) & 
$Q$: What's the shape of the earth?~~$A$: Spherical \newline 
\ck~: The earth is spherical. \newline 
\pk~: The earth is spherical. & $Acc, U_R$ \\
\midrule
\textit{Complementary} \newline \(\ck \land \pk~\) & 
$Q$: The shape of the planet \texttt{Z-man} was born on?~~$A$: Spherical \newline 
\ck~: \texttt{Z-man} was born on earth. \newline 
\pk~: The earth is spherical. & $Acc, U_R$ \\
\midrule
\textit{Conflicting} \newline \(\ck \oplus \pk~\) & 
$Q$: What's the shape of the earth?~~$A_{ck}$: Flat; $A_{pk}$: Spherical \newline 
\ck~: The earth is flat. \newline 
\pk~: The earth is spherical. & $M_R$ \\
\midrule
\textit{Irrelevant} \newline \(\ck \lor \pk~\) & 
$Q$: What's the shape of the earth?~~$A$: Spherical \newline 
\ck~: The earth orbits around the sun.   \newline 
\pk~: The earth is spherical. & $Acc, U_R$ \\
\bottomrule
\end{tabular}
\vspace{-0.5cm}
\label{tab:type_and_metrics}
\end{table*}

As shown in Table \ref{tab:type_and_metrics}, we design four reasoning types based on various relationships between \pk~and \ck. Without the loss of generality, we adopt Question Answering (QA) as the reasoning task, following prior work \citep{mallen-etal-2023-trust,cheng-etal-2024-call}. 
Formally, given an instruction, a question ${Q}$ and \ck, an LLM leverages \pk~and \ck~to reason the answer ${A}$. As shown in Equation \ref{reasoning_equation}, if $Q$ is posed, then \ck~operated with \pk~under the reasoning types (\textit{rt}) leads to the answer $A$. 
Note that for all settings, \pk~always holds true for Equation \ref{reasoning_equation}, meaning that we ensure the LLM has \pk~required for $Q$. In experiments, we vary \ck~based on our designed \textit{rt} as shown in Figure \ref{fig:main_example}. Different from \citealp{xie2024knowledgeconflict}, we only present  \ck~in the context, to assess the ability to utilize \pk~more naturally.


\begin{align}
&Q \rightarrow (\ck ~\textit{rt} ~\pk \rightarrow A) \textit{\quad where {rt}} \in \{\odot, \land, \oplus, \lor\} \textit{~and \pk~holds true.}   \label{reasoning_equation} 
\end{align}


\noindent \textbf{\textit{$\odot$ Supportive}} We have \ck~and \pk~convey the same information.
Our prior experiments show that models behave consistently with or without \ck~(as expected), or LLMs can effectively adopt \pk~in this case. Therefore, we do not report this type in main results. (Refer to Appendix \ref{sec:results_for_supportive} for statistics.)

\noindent\textbf{\textit{$\land$ Complementary}}\label{subsec:complementary_reasoning} This type typically requires both \pk~and \ck~(multi-hop)~for reasoning.
We maintain knowledge sufficiency by assuring \pk is known by LLM and \ck is new LLM. A key challenge is to establish meaningful connections between them. Previous study construct multi-hop questions by replacing entities with descriptions \citep{talmor-berant-2018-web,huang-etal-2023-markqa}, which does not ensure that \ck~is new to LLMs. To address this, we create \textit{entity-profiles} that relate to \pk. For the example in Table \ref{tab:type_and_metrics}, given \pk~``\textit{the earth is spherical}'', \ck~can be ``\textit{\texttt{Z-man} was born on earth}'', which is complementary to \pk~to answer ``\textit{The shape of the planet \texttt{Z-man} was born on}''.

\noindent\textbf{\textit{$\oplus$ Conflicting}} In this type, a fact in \ck~is fabricated by \textit{deletion, modification} or \textit{extension} from \pk. We can also create longer narratives around the fabricated fact. For the example in Figure \ref{fig:main_example}, by modification, we can change ``\textit{Michael Jordan is a Basketball Player}'' into ``\textit{Michael Jordan works as an ESPN basketball commentator for 20 years}''.
We ensure the question $Q$ require that exact fact to answer. 
We denote the answer that align with \ck~and \pk~as $A_{ck}$ and $A_{pk}$, respectively.

\noindent \textbf{\textit{$\lor$ Irrelevant}} For this type, we randomly sample irrelevant information as \ck, under certain constraints, \textit{e.g.,} \ck~for questions about \texttt{occupation} should still pertain to occupations. As  the goal is to test LLMs ability to leverage \pk,
we only consider the case when \pk~is necessary for $Q$ in main results. We discuss more on LLMs knowledge filtering ability in Appendix \ref{sec:ck_is_required}.

\subsection{Evaluation Metrics}

We design various metrics to assess LLMs ability to leverage \pk, as shown in Table \ref{tab:type_and_metrics}. Specifically:

\noindent $\bullet$ \textbf{Accuracy ($Acc$)} evaluates whether the output exactly matches the ground truth.
 For Supportive, Complementary, and Irrelevant reasoning types, each question $Q$ has a corresponding ground truth answer $A_{gt}$. We compute \textbf{$Acc$} to assess model performance for these reasoning types.
 
\noindent $\bullet$ \textbf{Memorization Ratio ($M_R$)} measures the extent to which the model adheres to \pk~for Conflicting Reasoning. It is calculated as $M_R = \frac{f_{pk}}{f_{pk}+f_{ck}}$, where $f_{pk}$ and $f_{ck}$ denotes the frequency of answer faithful to $A_{ck}$ and $A_{pk}$, respectively, following prior work \citep{longpre-etal-2021-entity}.

\noindent $\bullet$ \textbf{Unknown Ratio (${U_R}$)} measures the ratio of models generating unknown. 
On one hand, ${U_R}$  reveals models uncertainty to some extent. On the other hand, since we make sure \pk is required by $Q$, higher ${U_R}$ shows lower perception of \pk. We calculate $U_R$ for all reasoning types.

\subsection{Reasoning Instructions}
\label{sec:reasoning_instructions}
Recent post-training techniques, \textit{e.g.,} instruction-tuning \citep{chung2024scaling}, enable LLMs to fluently follow human instructions. To explore to what extent LLMs can leverage \pk, we design a series of progressively-enforced instructions to guide models behavior (detailed in Table \ref{tab:detailed_prompt_reasoning_instructions}).

\vspace{-0.2cm}
\begin{tcolorbox}[colback=black!15!white, colframe=black, title=Examples of Progressively-Enforced Reasoning Instructions]
    \vspace{-0.2cm}

    \textbf{\textit{Neutral}:} Combine the given information and your OWN knowledge to answer questions.
    \vspace{3pt}

    \textbf{\textit{Trust Yourself}}: 
     The given information is NOT SUFFICIENT, you should USE your OWN knowledge combined with the information to answer questions. \hfill(Complementary) \\
     If you think the given information is CONFLICTED with your knowledge, you should USE your OWN knowledge to answer questions.\hfill(Conflicting)

    \vspace{3pt}

    \textbf{\textit{Speak Out Loud}}: You must first OUTPUT your OWN knowledge about the question. If you think the given information is CONFLICTED with your knowledge, should USE your OWN knowledge to answer questions. \hfill(Conflicting)
    \vspace{-0.25cm}
    \label{tab:prompts}
\end{tcolorbox}

\noindent $\bullet$ \textbf{\textit{Neutral Instruction}} is natural and straightforward. We treat LLMs as a helpful assistant with knowledge to answer questions. Note that we indeed ask LLMs to use their own knowledge.

\noindent $\bullet$ \textbf{\textit{Trust Yourself Instruction}} proactively shows the insufficiency or asks LLMs to trust their own knowledge. Concerningly, such enforcing instruction is to test LLMs leverage of \pk, but not practical in real-world scenarios where the sufficiency and trustworthiness of knowledge are not ensured.

\noindent $\bullet$ \textbf{\textit{Speak Out Loud Instruction}} further pushes LLMs to explicitly utter its \pk~then trust themselves, mirroring the Chain-of-Thought reasoning \citep{wei2022chain}. This differs from \citealp{xie2024knowledgeconflict}, who separately constructs \pk~and \ck, and asks models to make a choice when \pk~and \ck~are both in context. Also, this enforcing instruction is not for practical use.

\subsection{Dataset}
\label{sec:dataset}
We introduce our dataset \textbf{\textsc{EchoQA}} to test how well LLMs echo their \pk~when encountering \ck.
Questions are in multi-choice format and provided an unknown option, following \citealp{xie2024knowledgeconflict}. The key step is to firstly obtain models \pk by questioning about a knowledge triple, where a correct answer indicates that the LLM possesses the relevant knowledge, as in the left of Figure \ref{fig:main_example}. Then, we construct \ck from \pk based on our designed reasoning types, as in the middle of Figure \ref{fig:main_example}. We conduct human evaluations on the quality of the constructed \ck and $Q$, showing that 97.6\% and 95.3\% of $Q$ and \ck~perfectly fit our reasoning types (Table \ref{tab:type_and_metrics}). Notably, our construction can adapt to any knowledge-intensive task. Specifically, \textsc{EchoQA} is adapted from the following datasets, covering scientific, factual and commonsense knowledge (more details in Appendix \ref{sec:details_of_data_construction}):

\begin{table}
  \centering
\caption{Number of examples for each LLM for Complementary/Conflicting/Irrelevant reasoning type. The difference between LLMs is due to their different possession of knowledge.}
\vspace{1pt}
  \begin{tabular}{lcccc}
    \toprule
    \textbf{Models} & \textbf{ALCUNA (\#)} & \textbf{ConflictQA (\#)} & \textbf{MuSiQue (\#)} & \textbf{OpenBookQA (\#)} \\ 
    \midrule
    OpenAI o1 & 537/512/1,119   & 1,993  & 611 & 488  \\ \midrule
    GPT-4o & 631/740/1851   & 3,001  & 525 & 476  \\ \midrule
    GPT-4o-mini & 599/852/1,811 & 1,801 & 675 & 458 \\ \midrule
    Llama 3.1-70B & 1,090/550/1,809 & 2,372 & 635 & 397 \\ \midrule
    Llama 3.1-8B & 905/873/1,953 & 2,299 & 393 & 351 \\ \midrule
    Qwen 2-7B& 993/346/1,397  & 2,143  & 452 & 415  \\ 
\bottomrule
  \end{tabular}
  \label{tab:number_of_filtered_questions}
\end{table}

\noindent \textbf{ALCUNA}~\citep{yin-etal-2023-alcuna} is a scientific (biological) dataset featuring artifact entities. The knowledge is structured in json format, obtained by adding, variation and dropping from existing knowledge, from which we construct conflict \ck. We adopt knowledge differentiation subset for Conflicting and Irrelevant Reasoning, and knowledge association subset for Complementary Reasoning, respectively. 
For fair comparison, we construct conflicting \ck under ``variation'' in Section \ref{sec:conflict_reasoning} and further discuss how the way of \ck construction affects LLMs awareness of \pk in Section \ref{sec:way_of_fabrication}.

 \noindent\textbf{ConflictQA}~\citep{xie2024knowledgeconflict} provides factual parametric evidence elicited from LLMs, conflicting evidence and corresponding answers by an LLM-based framework.
 We adopt the PopQA subset \citep{mallen-etal-2023-trust}, questioning about a single fact. Complementary \ck and questions are constructed by \textit{entity-profiles} with templates (Table \ref{tab:template_popQA}) for Complementary Reasoning. We adopt the original conflict knowledge and randomly sample several irrelevant evidences with the same relation type for Conflicting and Irrelevant Reasoning, respectively.

 \noindent \textbf{MuSiQue} \citep{trivedi2022musique} is a factual dataset featuring longer document and multi-step reasoning, with sub-questions and evidence. We firstly question LLMs with sub-questions to label supporting paragraphs as \pk and \ck. Then, we filter out examples with both \ck and \pk for Complementary Reasoning. We adopt {Llama 3.1-70B} (\textit{zero-shot}) to construct conflict \ck by modifying \pk~and generate related $Q$. Prompts are in Table \ref{tab:prompt_examples}. Other operations are similar to ConflictQA. 

 \noindent \textbf{OpenBookQA} \citep{mihaylov-etal-2018-openbook} is a commonsense multi-choice QA dataset. We elicit the commonsense \pk~by questioning LLMs with Chain-of-thought \citep{wei2022chain} and then filtering out ones with correct answers. We instruct {Llama 3.1-70B} with \textit{one-shot} demonstration to fabricate \ck and multi-hop $Q$ requiring both \pk~and \ck~for Complementary Reasoning. Please refer to Table \ref{tab:openbook_knowledge} for prompt examples. Other operations are similar to MuSiQue.


\subsection{Language Models}
\label{sec:language_models}

We include multiple LLMs spanning diverse parameter scales, \textit{i.e.,} two closed-source (OpenAI o1-preview \citep{openai2024o1preview}, GPT-4o and GPT-4o-mini \citep{openai2023gpt4o}) and three open-source (Llama 3.1-70B, Llama 3.1-8B \citep{dubey2024llama3herdmodels} and Qwen 2-7B \citep{yang2024qwen2}). These models have demonstrated strong performance in knowledge-intensive tasks and instruction-following capabilities, making them well-suited for our tasks. 
In \textsc{EchoQA}, we filter \pk for each LLM (as in the left of Figure \ref{fig:main_example}) to assess their ability to integrate \pk with \ck, statistics of examples are detailed in Table \ref{tab:number_of_filtered_questions}. For all LLMs, we experiment with Chain-of-Thought \citep{wei2022chain} for behavioral analysis and adopt \textit{zero-shot} setting to avoid the bias introduced by demonstrations. Other setups remain as default. Specific model API/checkpoints are listed in Appendix \ref{sec:details_exp_statics}.

\section{Experimental Results}
\label{sec:Experimental_Results}

\subsection{Complementary Reasoning}
\label{sec:complementary_reasoning}

\begin{figure*}[t]
\setlength{\abovecaptionskip}{-0.2cm}
\setlength{\belowcaptionskip}{-10pt}
  \includegraphics[width=\linewidth]{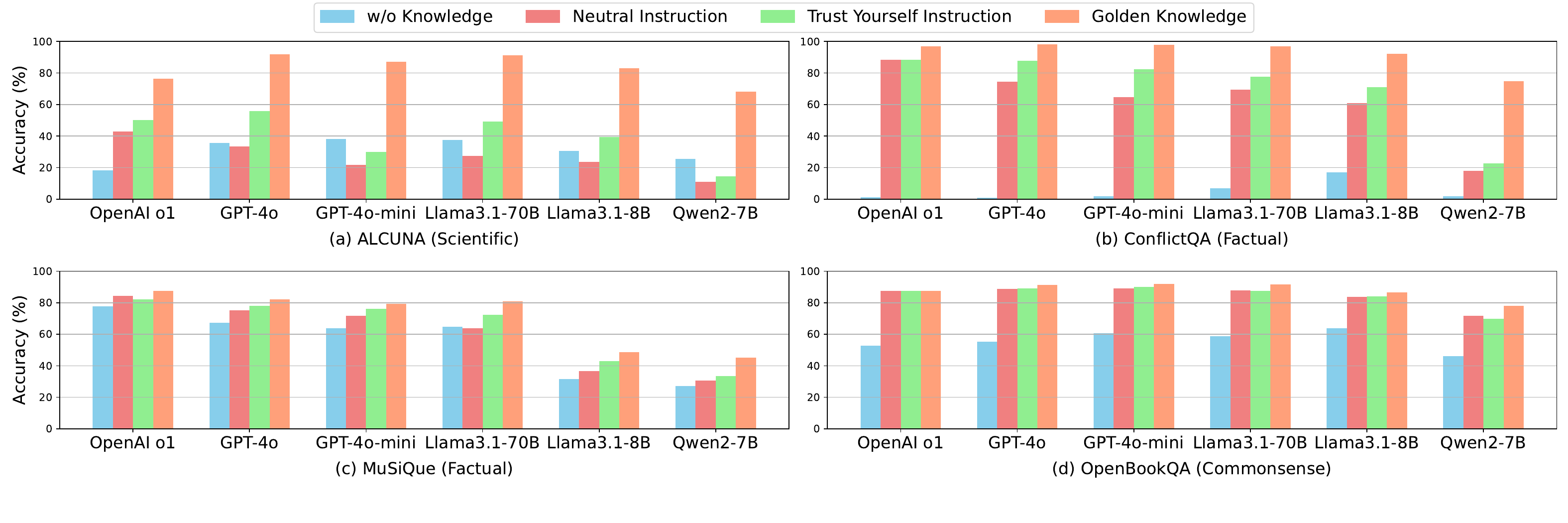}
  \caption {Accuracy for Complementary Reasoning. \textit{w/o} Knowledge and Golden Knowledge means no information and all the required information is given, respectively. The upward trend shows that LLMs suppress \pk~even with complementary \ck, comparing to the orange bar.}
  \label{fig:complementary_reasoning}
\end{figure*}

\textbf{Although provided with complementary context, LLMs leverage of their own knowledge remains inhibited.} Figure \ref{fig:complementary_reasoning} illustrates the Accuracy of LLMs for questions requiring both their parametric knowledge (\pk) and the contextual knowledge (\ck). Table \ref{tab:ur_for_complementary} shows the Unknown Ratio $U_R$ on ALCUNA (more details in Appendix \ref{sec:results_for_unknown_ratio}). We also experiment with no knowledge and all required knowledge, \textit{i.e.,} \ck~and elicited \pk, in the context as the lower and upper bound (blue and orange bars), respectively. The following observations can be drawn:

\begin{wraptable}{r}{0.47\textwidth}
    \setlength{\abovecaptionskip}{-0.45cm} 
    \setlength{\belowcaptionskip}{1pt} 
    \setlength{\tabcolsep}{4pt}  
    \caption{Unknown Ratio $U_R$ (\%) for Complementary Reasoning on ALCUNA. \textit{w/o} K, Gold K means none and golden knowledge is given, respectively. NI, TYI denotes Neutral and Trust Yourself Instruction, respectively. The \textbf{bold} denotes the highest $U_R$ in each row.}
    \centering
    \small
    \begin{tabular}{lcccc}
        \toprule
        \textbf{Models} & \textbf{\textit{w/o} K} & \textbf{NI} & \textbf{TYI} & \textbf{Gold K} \\ 
        \midrule
           \rowcolor{gray!20} \multicolumn{5}{c}{\textit{Closed-source LLMs}} \\ 
        OpenAI o1 & \textbf{65.78} & 46.12 & 35.43 & 20.63 \\ 
        GPT-4o & 36.45 & \textbf{59.90} & 26.94 & 8.08 \\ 
        GPT-4o-mini & 46.12 & \textbf{76.13} & {68.28} & 12.52 \\ 
        \midrule
        \rowcolor{gray!20} \multicolumn{5}{c}{\textit{Open-source LLMs}} \\ 
        Llama 3.1-70B & 23.89 & \textbf{62.72} & 23.88 & 0.08 \\ 
        Llama 3.1-8B & 18.12 & \textbf{58.50} & 34.35 & 12.70 \\ 
        Qwen 2-7B & 40.28 & \textbf{81.26} & 73.62 & 28.60 \\ 
        \bottomrule
    \end{tabular}
    \label{tab:ur_for_complementary}
    \vspace{-0.2cm}
\end{wraptable}

First, \textit{LLMs cannot fully leverage \pk, even though the given \ck~with \pk~is sufficient for the question}. All tested LLMs across all knowledge types demonstrate an upward trend in Accuracy, with stronger models exhibiting higher absolute performance. 
With all required knowledge in context, LLMs perform effectively (orange bar). However, the performance drops dramatically with only \ck in context (red/green bar). Figure \ref{fig:complementary_reasoning} (a) (scientific knowledge) shows that the introduction of complementary \ck even degrades the performance for most tested LLMs, compared to no knowledge given. We scrutinize 200 samples and find that the presence of \ck hinders LLMs from engaging in deeper reasoning, \textit{e.g.,} inferring entity properties by names, leading to an over-reliance on context. We further discuss the phenomenon in Section \ref{sec:factors_affecting}. 


Second, \textit{with complementary \ck, LLMs show even increased uncertainty.} Table \ref{tab:ur_for_complementary} indicates that the complementary scientific \ck~``confuses'' most tested LLMs significantly, doubling the $U_R$ for Llama and Qwen models, comparing ``NI'' with ``\textit{w/o} K'' column. This implies that LLMs depend excessively on the relevant context to seek answers, suppressing their own \pk even though \pk is required for the question.



\textbf{Instruction-following aids in recalling some memories, but still way off.}
By asking LLMs to adopt \pk~in a progressively enforced tone, our reasoning instructions (Section \ref{sec:reasoning_instructions}) significantly bring up the performance, demonstrating the strong instruction-following capabilities of LLMs and showing that instructions can modulate perception of knowledge to some extent. However, even when explicitly guided, LLMs still cannot fully recall the necessary \pk~they already have, leaving a large margin compared to when all knowledge is provided in the context. This highlights the need for dedicate prompt or framework design for better integration of different sources of knowledge.

\noindent\textbf{Commonsense knowledge is easier to remember, while the scientific and factual is harder.}
Comparing different types of knowledge in Figure \ref{fig:complementary_reasoning}, we have the following observations:

\noindent First, LLMs exhibit nearly doubled accuracy in commonsense relative to scientific knowledge (red and green bar Figure \ref{fig:complementary_reasoning} (a)(d)). We speculate the higher recall of specific \pk is likely because such information appears more frequently the in training corpus (more discussion in Section \ref{sec:factors_affecting}). 

\noindent Second, LLMs can hardly answer without knowledge for ConflictQA (blue bar in Figure \ref{fig:complementary_reasoning}(b)). Different from other datasets, to fit the original question, we fabricate entity profiles and questions by templates in Table \ref{tab:template_popQA}, \textit{e.g.,} ``\textit{What's the occupation of the best friend of PersonA in high school?}'', leaving little room for inference. This inversely suggests LLMs reasonable reasoning ability.

Moreover, we observe that the upper bound for MuSiQue is considerably lower, especially for weaker models. However, the upward trend in performance persists, suggesting that while multi-step questions challenge LLMs reasoning ability, they may not significantly impair the recall of \pk.

\subsection{Conflicting Reasoning}
\label{sec:conflict_reasoning}

\begin{figure*}[t]

\setlength{\abovecaptionskip}{-5pt}
\setlength{\belowcaptionskip}{-10pt}

  \includegraphics[width=\linewidth]{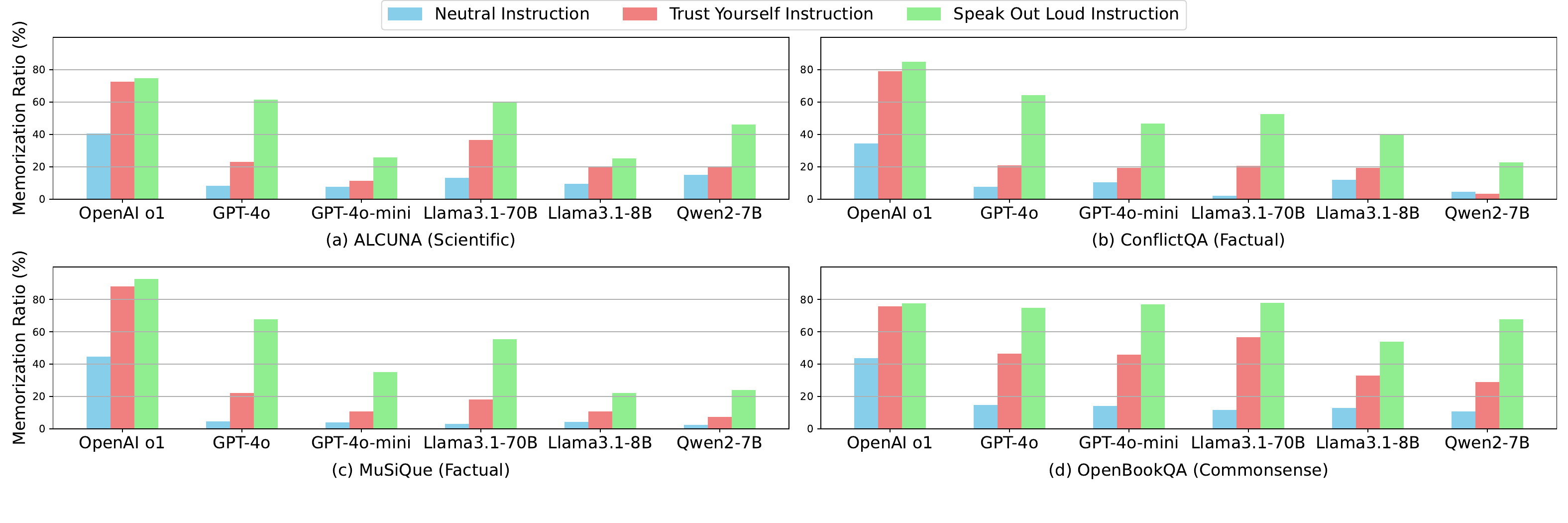}
  \caption {Memorization Ratio for Conflicting Reasoning. LLMs rarely trust themselves (\pk) when faced with conflicting \ck, though instructions modulate their preference to some extent.}
  \label{fig:conflicting_reasoning}
\end{figure*}

\textbf{LLMs rarely trust themselves when faced with conflicting context.} Figure \ref{fig:conflicting_reasoning} depicts the ratio of LLMs relying on their parametric knowledge (\pk) when presented with conflicting contextual knowledge (\ck). Under Neutral Instruction, other than the strong OpenAI o1, few LLMs lean on themselves for more than 10\% cases across all datasets. Note that the unknown ratio here is relatively low (more in Appendix \ref{sec:results_for_unknown_ratio}), making it reasonable to draw conclusions from Figure \ref{fig:conflicting_reasoning}. It is demonstrated that \textit{LLMs are faithful to the conflict context across all types of knowledge}, including commonsense, echoing findings in prior work \citep{xie2024knowledgeconflict}. This observation underscores limitations in LLMs confidence and usage on their own knowledge when confronted with disinformation, warranting dedicated content filtering and careful safety design for LLM-based systems.

\textbf{Instructions significantly influence LLMs preferences, but still struggle to fully recall their memory.}
We further explore the extent to which LLMs can recall \pk~when faced with conflicting \ck. As we instruct LLMs to trust themselves in a progressively  more enforced tone, the Memorization Ratio ($M_R$) consistently goes up in all datasets, with stronger model exhibiting relatively higher $M_R$ (Figure \ref{fig:conflicting_reasoning}). Compared to Neutral Instruction, the Trust Yourself Instruction almost doubles the $M_R$, with the Speak Out Loud Instruction further boosting the $M_R$. However, despite being instructed to trust themselves, few of $M_R$ exceed 60\% for scientific and factual knowledge, indicating that LLMs still cannot fully recall their internal knowledge when the context is in conflict. Again, o1 exhibits more confidence than others, especially when told to ``Trust Yourself'', but there is still space for improvement. This also highlights the vulnerability of LLMs prone to disinformation.

\textbf{LLMs exhibit greater resistance to conflicts in commonsense knowledge.}
Figure \ref{fig:conflicting_reasoning}(d) shows that LLMs can recall commonsense knowledge relatively better, with our instructions leads to stronger enhancement than others. Specifically, 
most tested LLMs are confident against conflicting \ck in over 60\% cases with the ``Speak Out Loud Instruction'' (green bar). Since commonsense facts are less likely to become outdated or change over time, it is crucial that LLMs stand on themselves when encountering conflicting information. Although not perfect, this offers promising insights into preventing attacks. 
In contrast, while the upward trend still holds, most LLMs are less likely to trust their own factual knowledge, often succumbing to conflicting \ck~(Figure \ref{fig:conflicting_reasoning}(b)(c)), highlighting the importance of knowledge authenticity in LLM-based systems.

\subsection{Irrelevant Reasoning}
\label{sec:irre_reasoning}
\begin{figure*}[t]

\setlength{\abovecaptionskip}{-5pt}
\setlength{\belowcaptionskip}{-10pt}
  \includegraphics[width=\linewidth]{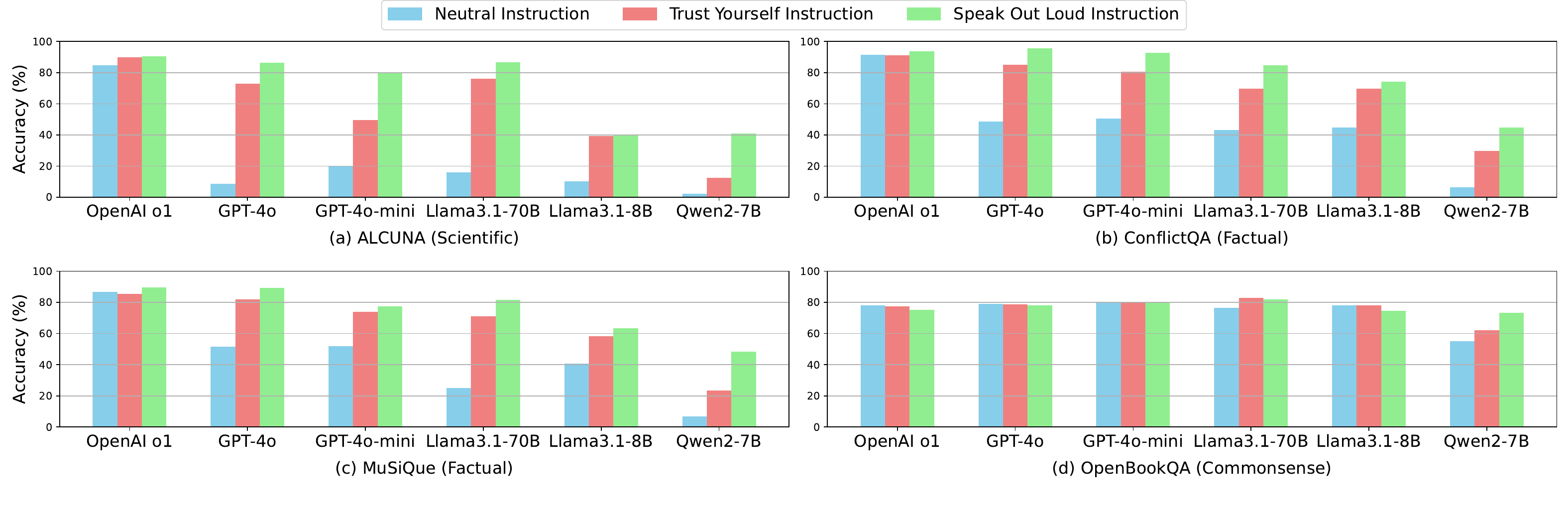}
  \caption {Accuracy for Irrelevant Reasoning, showing that LLMs relying on \ck~even though it is irrelevant and that instructions can substantially modulate their leverage of knowledge.}
  \label{fig:irrelevant_reasoning}
\end{figure*}

\textbf{Although grasping the key to the knowledge, LLMs still seek answers in the irrelevant context.}
Figure \ref{fig:irrelevant_reasoning} illustrates the Accuracy for questions requiring LLMs knowledge (\pk), when provided with one to four irrelevant documents in context (\ck). Table \ref{tab:ur_for_conflict} shows the Unknown Ratio ($U_R$) on ConflictQA (refer to Appendix \ref{sec:results_for_unknown_ratio} for more details). Note that we indeed ask LLMs to adopt their own knowledge in all instructions.
Under Neutral Instruction, although knowing the answers and the provided information is irrelevant, most LLMs, other than o1, tend to search for answers solely within the \ck, particularly in the case of scientific knowledge (blue bar). This indicates \textit{the difficulty LLMs face in disentangling \pk~from \ck~during reasoning}. Again, commonsense knowledge is more easily recalled, with GPT and Llama achieving up to 80\% accuracy using only Neutral Instruction. We further discuss LLMs filtering ability in Appendix \ref{sec:ck_is_required}).

\begin{wraptable}{r}{0.44\textwidth}
    \setlength{\abovecaptionskip}{-12pt} 
    \setlength{\belowcaptionskip}{1pt} 
    \setlength{\tabcolsep}{5.9pt}  

    \centering
    \caption{Unknown Ratio $U_R$ (\%) for Irrelevant Reasoning on ConflictQA. NI, TYI, SOLI denotes Neutral, Trust Yourself and Speak Out Loud Instruction, respectively. The \textbf{bold} denotes highest $U_R$ in each row.}
    \begin{tabular}{lccc}
        \toprule
        \textbf{Models} &  \textbf{NI} & \textbf{TYI} & \textbf{SOLI} \\ 
        \hline
        \rowcolor{gray!20} 
        \multicolumn{4}{c}{\textit{Closed-source LLMs}} \\ 
        OpenAI o1 & \textbf{6.12} & \textbf{6.12} & 0.98  \\ 
        GPT-4o & \textbf{50.38} & 13.46 & 0.53  \\ 
        GPT-4o-mini & \textbf{48.42} & 17.81 & 0.80  \\ 
        \hline
        \rowcolor{gray!20} 
        \multicolumn{4}{c}{\textit{Open-source LLMs}} \\ 
        Llama3.1-70B & \textbf{55.14} & 27.36 & 2.33  \\ 
        Llama3.1-8B & \textbf{50.41} & 23.88 & 4.83 \\ 
        Qwen2-7B & \textbf{80.59} & 41.82 & 4.57 \\ 
        \bottomrule
    \end{tabular}
    \vspace{-10pt}
    \label{tab:ur_for_conflict}
\end{wraptable}
We also observe that when the given information is irrelevant, other than o1, LLMs exhibit high uncertainty, with more than 50\% Unknown Ratio for most tested models. With our progressively-enforced Reasoning Instructions, the uncertainty decreases both gradually and substantially, dropping to below 5\% for all tested models on ConflictQA (Table \ref{tab:ur_for_conflict}). Such result also suggests higher possibility of models leverage of \pk. Moreover, the OpenAI o1 shows extraordinary ability echoing \pk when \ck is irrelevant, showing close results at different levels of instructions. This indicates the intricate design behind o1 system and sheds lights on future improvements.

\textbf{LLMs echo their knowledge relatively better when the context is irrelevant.}
Comparing the absolute performance with Conflicting and Complementary Reasoning (Figure \ref{fig:complementary_reasoning} and \ref{fig:conflicting_reasoning}), LLMs are able to better leverage \pk~for reasoning when \ck is irrelevant. With our progressively-enforced Reasoning Instructions, LLMs further improve their memory recall by a large margin (Accuracy reaches 80\% for most tested models Figure \ref{fig:irrelevant_reasoning}) and become more certain about their answers (Unknown Ratio drops drastically in Table \ref{tab:ur_for_conflict}). The observation, alongside the results for Complementary and Conflicting Reasoning, indicates the possibility of leveraging \pk~for reasoning. However, any knowledge presented in context tends to capture their attention to some extent. In other words, \textit{the relationships between \pk~and \ck~greatly influence LLMs awareness and leverage of their knowledge.}

\section{Discussion}
\label{sec:discussion}

\subsection{Models Leverage of Parametric Knowledge Given Contextual Knowledge}

In previous sections, we comprehensively investigate LLMs behavior in the dynamic integration of parametric knowledge (\pk) and contextual knowledge (\ck). Specifically:

\noindent $\bullet$ We identify four reasoning types based on various relationships between \pk~and \ck. To test models ability to leverage \pk~when encountering \ck, we introduce a new dataset \textsc{EchoQA}, spanning scientific, factual and commonsense knowledge (Section \ref{sec:experimental_setup}). 

\noindent $\bullet$ Our observations show that while LLMs possess the required knowledge, the presence of contextual information can significantly suppress their ability to leverage \pk, regardless of the model, reasoning type, or knowledge type (Section \ref{sec:Experimental_Results}).

\noindent $\bullet$ We also observe that certain types of knowledge are easier to recall. LLMs can better utilize their commonsense knowledge or when the context is irrelevant, compared to other types. This sheds lights on the improvement of integration of \pk~and \ck~to enhance reasoning performance.

\noindent $\bullet$ To improve the utilization of \pk, we designed a range of reasoning instructions. Results demonstrate excellent instruction-following capabilities of LLMs. When instructed to ``Speak Out Loud'' or ``Trust Yourself'', LLMs substantially improve their memory recall. However, a significant gap still remains in fully leveraging \pk~when \ck~is present (Figures \ref{fig:complementary_reasoning}, \ref{fig:conflicting_reasoning}, \ref{fig:irrelevant_reasoning}).

    

\begin{figure*}[t]
\setlength{\abovecaptionskip}{-5pt}
\setlength{\belowcaptionskip}{-15pt}
 \centering
  \includegraphics[width=0.95\linewidth]{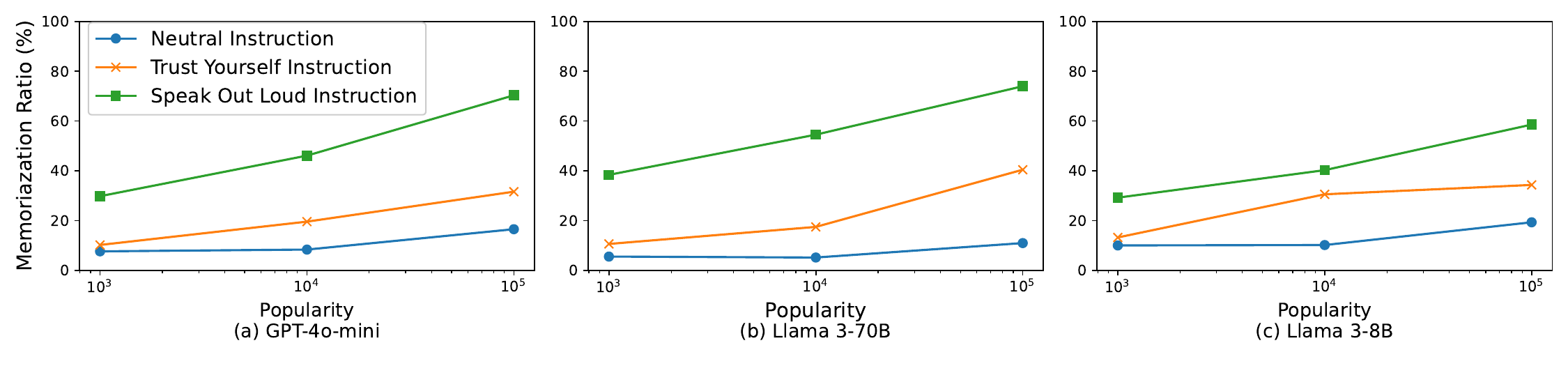}
  \caption {Memorization Ratio on ConflictQA across \textit{popularity} categories by representative models, showing that LLMs recall their popular knowledge better.}
  \label{fig:ppl_graph}
\end{figure*}

\subsection{Factors Affecting Models Leverage of Parametric Knowledge}\label{sec:factors_affecting}
In previous findings, we show that the awareness and utilization of \pk~can be influenced by instructions, knowledge type, and the intricate relations with \ck. Here, we discuss additional factors:

\noindent \textbf{LLMs can recall their popular knowledge more effectively.} Inspired by studies showing that LLMs lean on more popular entities, \textit{i.e.,} monthly associated Wikipedia page views \citep{mallen-etal-2023-trust,xie2024knowledgeconflict}, we evaluate representative LLMs on ConflictQA with Conflicting Reasoning \textit{w.r.t,} popularity of the topic entity in the question, as illustrated in Figure \ref{fig:ppl_graph}. Intuitively, LLMs can recall memories better for more popular questions. This upward trend holds across all LLMs and our reasoning instructions. Specifically, GPT-4o-mini and Llama 3.1-70B achieve nearly 80\% Memorization Ratio for the most popular questions, almost doubling that of the least popular ones. This observation might also explain why LLMs remember better for commonsense knowledge than others, indicating the reason behind is likely the knowledge frequency in training corpus.

\textbf{Introduction of knowledge relevant to the question may draw overly excessive attention to the context.} As mentioned in Section \ref{sec:complementary_reasoning} (Figure \ref{fig:complementary_reasoning}(a)), counter-intuitively, even though the given \ck~with \pk~is sufficient to answer the question, performance can still degrade compared to direct questioning. We manually examine 200 random samples generated by Llama 3.1-70B with Neutral Instruction on ALCUNA where LLMs were able to answer on their own but failed when complementary information was introduced. Note that we do ask LLMs to use their own knowledge in the instruction (Section \ref{sec:reasoning_instructions}). We discover that in 93.5\% of cases, LLMs relied on the context and answer choices and even claimed the insufficiency of knowledge (Refer to Appendix \ref{sec:error_analysis}, Table \ref{tab:case_study_2} and \ref{tab:case_study_1}). This suggests that  \textit{LLMs tend to ignore their own knowledge, inhibit deeper thinking and depend excessively on the context}. Although our ``Trust Yourself Instruction'' improves performance, it is still far from expectation, indicating that \textit{current LLMs struggle with integrating \pk~and \ck.}

We speculate that the phenomenon may stem from the post-training process \citep{chung2024scaling}. LLMs are fine-tuned with instruction, question and optional context in a fashion shown in Equation \ref{eqa:training}. This training approach encourages LLMs to answer based on either solely the context (\ck) or their internal knowledge (\pk) separately, rather than integrating both. We hypothesize that this training objective is the reason behind LLMs struggles in effectively combining \pk~and \ck.

\vspace{-0.4cm}
\begin{equation}
    \textit{Instruction;~Context} \text{(Optional)};~\textit{Question} \rightarrow ~\textit{Answer} \\ \label{eqa:training}
\end{equation}

\begin{figure*}[t]
\setlength{\abovecaptionskip}{-0.1cm}
\setlength{\belowcaptionskip}{-0.4cm}
 \centering
  \includegraphics[width=0.95\linewidth]{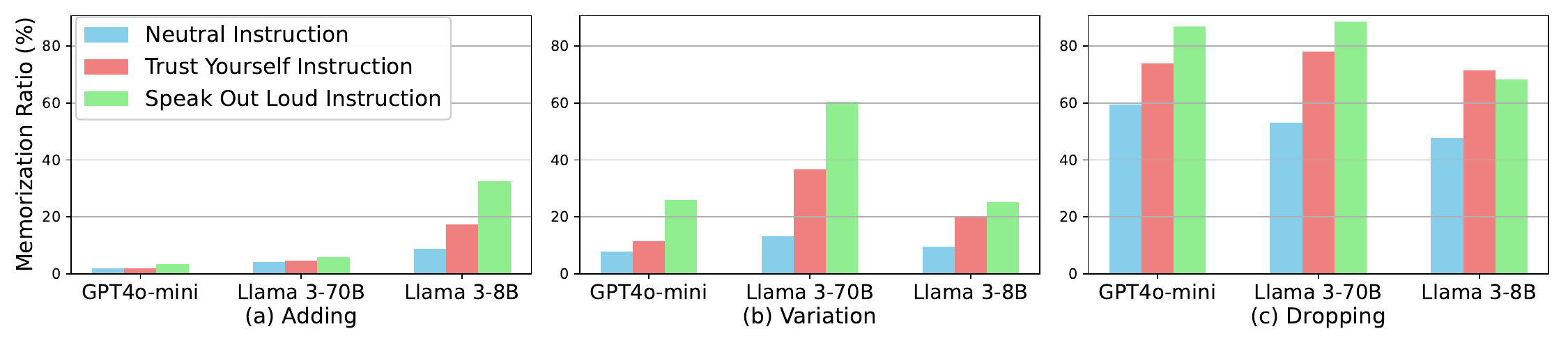}
  \caption {Memorization Ratio for Conflicting Reasoning under different ways of fabricating \ck~on
 ALCUNA. LLMs recall their knowledge way better with dropping than variation and adding.}
  \label{fig:alcuna_conflict}
\end{figure*}

\noindent \textbf{The method of new knowledge fabrication impacts LLMs recall of memory.} \label{sec:way_of_fabrication}
 As mentioned in Section \ref{sec:experimental_setup}, conflicting facts (\ck) can be introduced through various methods: \textit{adding}, \textit{variation}, and \textit{dropping}.
For fair comparison, we report results under ``variation'' in Figure \ref{fig:conflicting_reasoning}, since all datasets introduce conflicts by default through this method.
Here, we further investigate how different fabrication methods affects LLMs awareness and leverage of \pk, by experimenting representative LLMs on ALCUNA in Conflicting Reasoning. Figure \ref{fig:alcuna_conflict} provides the following insights:

First, \textit{a significant gap exists between different fabrication methods}, although the overall trend observed in previous experiments remains consistent. This highlights that LLMs awareness of \pk~is strongly related to the how \ck is fabricated from (or related to) their original \pk.
 
Second, \textit{when new properties are added to an existing entity, LLMs are highly receptive to the new information}, exhibiting mostly lower than 10\% Memorization Ratio. This raises safety concerns, as LLMs may be convinced of the addition of disinformation to known entities, \textit{e.g.,} fake news. 

Third, \textit{dropping of known facts does less harms in LLMs memory recall}. Compared with the other methods, \ck constructed by dropping provides some information about the entity irrelevant to the question. In Figure \ref{fig:alcuna_conflict}(c), with only Neutral Instruction, LLMs can leverage \pk in up to 60\% of cases, echoing results in Irrelevant Reasoning (Section \ref{sec:irre_reasoning}) where LLMs recall their memories better than other reasoning types. However, they are still far from fully echoing their \pk. We further discuss models ability to filter relevant information in Appendix \ref{sec:ck_is_required}.

\subsection{How to Better Leverage Parametric Knowledge Given the Context}

As demonstrated above, LLMs struggle to integrate their parametric knowledge (\pk) with contextual knowledge (\ck). However, this ability is crucial for both artificial and human intelligence. We have designed instructions to elicit \pk~to some extent (Section \ref{sec:reasoning_instructions}). Although effective, these instructions are progressively-enforced and may not be natural enough for practical applications. The key challenge is to enable LLMs to naturally learn how to balance the two sources of knowledge.
A promising direction is to develop an agent framework where LLMs assume a controlled amount of responsibility. By separating the process into two steps: first recalling the required knowledge, then reasoning, LLMs may avoid distraction from the context. Another potential solution is to fundamentally integrate \ck~and \pk~in post-training (Equation \ref{eqa:training}), which we leave for future work.

\section{Conclusion}
In this work, we comprehensively investigate how well LLMs can leverage their parametric knowledge (\pk) in the presence of contextual knowledge (\ck). We formulate four \textit{reasoning types}: \textit{Supportive, Complementary, Conflicting and Irrelevant}, based on diverse relationships between \ck~and \pk. To facilitate the study, we introduce a new dataset \textsc{EchoQA} spanning scientific, factual and commonsense knowledge for multiple LLMs. We design a series of \textit{reasoning instructions} with progressively more enforced tones to showcase LLMs leverage of \pk. We find that LLMs consistently struggle to fully leverage their \pk~when \ck~is present under different reasoning types. While the tailored reasoning instructions substantially improve their awareness of \pk, they still fall short of expectations. Furthermore, we show that LLMs awareness of their \pk~can be affected by the relation between \ck and \pk, instruction, entity popularity and methods of fabricating \ck, showing the vulnerability when faced with external knowledge. We hope that our work provides valuable resources and insights to guide future explorations in understanding LLM-based systems. We will release our code and dataset to facilitate future research at \href{https://github.com/sitaocheng/Knowledge_Interplay}{https://github.com/sitaocheng/Knowledge\_Interplay}.


\bibliography{iclr2025_conference}
\bibliographystyle{iclr2025_conference}

\newpage

\appendix
\section*{Limitations}
Our study reals a key vulnerability in current LLMs: they struggle to fully leverage their parametric knowledge when contextual knowledge is present. However, there are some limitations to our work. First, we base our study primarily on contemporary LLMs with strong reasoning and instruction-following abilities. While we find evidence that this issue can be partially mitigated through dedicated prompt or framework design, we argue that the ability to effectively integrate \pk with \ck should be inherent to the base model. We aim for improvements in this area in future work.
Second, while we consistently observe our findings across various reasoning types, we did not focus on making the data more challenging for broader coverage. For instance, conflicting knowledge could occur at the entity, sentence, or document level. A deeper exploring on such distinctions is left for future research.
Third, real-world applications are likely more complex than the reasoning types we have designed. Although we categorize the interactions between \pk and \ck into different reasoning types, there may be more nuanced combinations of these relationships in real-world scenarios. We hope our findings raise awareness of the safety concerns surrounding the trustworthiness and reliability of LLM-based systems.

\section*{Appendix}
In the following sections, we detail on the following aspects:
\begin{itemize}
    \item Appendix \ref{sec:details_of_data_construction}: Details of Data Constructions
    \item Appendix \ref{sec:details_exp_statics}: Details of Experimental Statistics
    \item Appendix \ref{sec:prompt_design}: Prompts Examples
\end{itemize}
\section{Elaboration on Data Construction}
\label{sec:details_of_data_construction}
As illustrated in Section \ref{sec:dataset}, our proposed dataset \textbf{\textsc{EchoQA}} aims to test LLMs awareness and leverage of parametric knowledge (\pk) given contextual knowledge (\ck). EchoQA is adapted from existing datasets, the construction procedure of which can fit into any knowledge-intensive task. The key step is to obtain LLMs \pk then construct \ck and answer choices to fit our reasoning types, as in Figure \ref{fig:main_example}. By default, the answer choice includes the answer sticking to \ck~and~\pk~for Conflicting Reasoning or the ground truth for other reasoning types, a randomly-sampled option with the same type, and an unknown option. We conduct human evaluation for the generated questions and contextual knowledge to ensure the meaningfulness, knowledge faithfulness and fluency. Examples for each dataset is shown in Table \ref{tab:complementary_dataset_exp}, \ref{tab:conflict_dataset_examples}, \ref{tab:irrelevant_dataset_exp}, \textit{w.r.t,} our reasoning types. Here we show more details.

\textbf{ALCUNA} \citep{yin-etal-2023-alcuna} originally provides new knowledge. To obtain \pk, we adopt a multi-choice question about the fact required by the original question. For example, \textit{``Does cat prey on fish? A. Yes, B. No, C. Unknown''} or \textit{``Cat prey on what? A. Fish, B. Tiger, C. Unknown''}. The exact fact for the question is provided by the dataset. The knowledge association subset naturally fits our Complementary Reasoning Type. Questions for other reasoning types are sampled from the knowledge differentiation subset. For Conflicting Reasoning, we regard the new knowledge adapted from the old one as a conflict fact by changing the new entity name back. For example, given \textit{``Dogcat prey on rabbit''}, we change it back to \textit{``Cat prey on rabbit''} as a conflict fact. The entity name in the question also need to change accordingly. For Irrelevant Reasoning, we randomly sample 1-4 new entities profiles provided as \ck.

\textbf{ConflictQA} \citep{xie2024knowledgeconflict} originally provides conflicting factual knowledge based on PopQA \citep{mallen-etal-2023-trust} and StrategyQA \citep{geva-etal-2021-aristotle}. We adopt the PopQA subset, with the \textit{popularity} of a question, counted by monthly Wikipedia page views associated with entities mentioned in the question. The key challenge is to construct new knowledge never seen by the LLMs for Complementary Reasoning. Following \citealp{mallen-etal-2023-trust}, we adopt templates for each relationship (Table \ref{tab:template_popQA}). With the template, we obtain \ck~and construct new multi-hop questions. For example, for question \textit{``What's the occupation of Michael Jordan''}, we adopt template to create a new entity ``PersonA'' with profile \textit{``Michael Jordan is the famous friend of PersonA''}. And we construct the question \textit{``What's the occupation of the famous friend of PersonA''}. For Irrelevant Reasoning, we randomly sample provided knowledge with the same relationship.

\textbf{MuSiQue} \citep{trivedi2022musique} provides all supporting paragraphs, from which we can query LLMs to obtain the \pk~and \ck, in a manner similar to ALCUNA. The sub-questions and documents can be directly used for Complementary Reasoning. We also adopt the answers of sub-questions as answer choices for Complementary Reasoning. The key challenge for this dataset is to create conflicting knowledge. We prompt Llama 3.1-70B by zero-shot to generate conflicting knowledge and a related question, based on the filtered \pk, for Conflicting Reasoning. The prompt is shown in Table \ref{tab:prompt_examples}. The constructed questions are also used for Irrelevant Reasoning, where we sample irrelevant documents from the provided unused paragraphs to test factual knowledge with a longer context.

\textbf{OpenBookQA} \citep{mihaylov-etal-2018-openbook} provides some scientific facts. To obtain \ck~and \pk, we query LLMs to answer the question in a Chain-of-Thought \citep{wei2022chain} manner (Refer to Table \ref{tab:openbook_knowledge} for prompt examples). We thereby obtain the \pk by filtering the thoughts from items with the correct answers. Then, we prompt Llama 3.1-70B with one demonstration to generate complementary knowledge and questions (Refer to Table \ref{tab:openbook_knowledge} for prompt examples). The Conflicting and Irrelevant Reasoning data construction is similar a MuSiQue.

\begin{table}[t]
    \caption{Templates to connect new knowledge to parametric knowledge for PopQA. [subj] denotes subject entity in the new question. For example, for question \textit{``What's the occupation of Michael Jordan''}, we adopt template to create a new entity with profile \textit{``Michael Jordan is the famous friend of PersonA''}. And we change the question to \textit{``What's the occupation of the famous friend of PersonA''}.}
    \vspace{3pt}
    \centering
    \begin{tabular}{>{\raggedright}m{3cm}|>{\raggedright\arraybackslash}m{8cm}}
        \toprule
        \textbf{Relationship} & \textbf{Template} \\ \midrule
        Occupation & 
        the person who frequently collaborates with [subj]\newline the best friend of [subj]\newline the famous friend of [subj] \\ \midrule
        Place of Birth & 
        the renowned figure who is a friend of [subj]\newline the best friend of [subj]\newline the person who frequently collaborates with [subj] \\ \midrule
        Genre & 
        the influential entity previously related with [subj] \\ \midrule
        Father & 
        the renowned figure with [subj]\newline the famous person known by [subj]\newline the best friend of [subj] \\ \midrule
        Country & 
        the renowned entity similar to [subj]\newline the notable figure similar to [subj] \\ \midrule
        Producer & 
        the renowned product similar to [subj]\newline the famous product similar to [subj]\newline the prominent product like [subj] \\ \midrule
        Director & 
        the renowned film similar to [subj]\newline the famous film similar to [subj]\newline the renowned movie similar to [subj] \\ \midrule
        Capital of & 
        the famous place near [subj]\newline the renowned place near [subj]\newline the prominent place similar to [subj] \\ \midrule
        Screenwriter & 
        the renowned screenplay similar to [subj]\newline the famous TV shows similar to [subj]\newline the prominent show like [subj] \\ \midrule
        Composer & 
        the work similar to [subj]\newline the prominent work similar to [subj]\newline the influential work similar to [subj] \\ \midrule
        Color & 
        the notable entity similar to [subj]\newline the prominent entity similar to [subj] \\ \midrule
        Religion & 
        the best friend of [subj]\newline the prominent friend of [subj] \\ \midrule
        Sport & 
        the prominent entity related to [subj]\newline the famous entity similar to [subj] \\ \midrule
        Author & 
        the prominent work similar to [subj]\newline the influential work similar to [subj] \\ \midrule
        Mother & 
        the renowned figure with [subj]\newline the famous person known by [subj]\newline the best friend of [subj] \\ \midrule
        Capital & 
        the important place near [subj]\newline the big place near [subj]\newline the neighbor place of [subj] \\ \bottomrule
    \end{tabular}
    \label{tab:template_popQA}
\end{table}

\section{Elaboration on Experiments}
\label{sec:details_exp_statics}

\subsection{Language Model API/Checkpoints}

This section elaborates on checkpoints we adopt for experiments. All open-source models are available on the Hugging Face platform. For Llama 3.1, we use ``meta-llama/Llama-3.1-(8, 70)B-Instruct'', respectively. For Qwen 2, we use ``Qwen/Qwen2-7B-Instruct''. We use OpenAI o1-preview, GPT-4o and GPT-4o-mini from OpenAI platform (https://platform.openai.com). 

\subsection{Results for Supportive Reasoning}
\label{sec:results_for_supportive}

\begin{wraptable}{r}{0.45\linewidth} 
    \setlength{\abovecaptionskip}{-0.35cm}
    \caption{Accuracy for Supportive Reasoning, with single- and multi-hop questions on ALCUNA.}
    \centering
    \small
    \vspace{3pt}
    \begin{tabular}{lcc}
        \toprule
        \textbf{Model} & \textbf{Single-hop}  & \textbf{Multi-hop}  \\
        \midrule
        GPT-4o-mini & 98.97 & 87.15  \\
        Llama 3.1-70B & 97.80 & 91.30  \\
        Llama 3.1-8B & 93.29 & 82.87  \\
        \bottomrule
    \end{tabular}
    \label{tab:supportive_reasoning}
\vspace{-3pt}
\end{wraptable}

Here we report the results for Supportive Reasoning on ALCUNA with representative LLMs. We do not include it in main experiments for two reasons: 1) Intuitively, models perform quite well in this type. As we make share the \pk is held by LLMs and \ck represents the same knowledge, the result is straightforward; 2) More importantly, it is hard to tell whether LLMs use \ck~or \pk~for reasoning when \ck~is present. We test questions require single and multiple facts for reasoning, as shown in Table \ref{tab:supportive_reasoning}. It is demonstrated that LLMs performance quite well with this type, reaching our expectation. However, based on our previous conclusions that LLMs might excessively rely on the context for reasoning, so they still may not perfectly leverage their own knowledge.

\subsection{Analysis of Models Uncertainty}
\label{sec:results_for_unknown_ratio}

\begin{figure*}[t]
\setlength{\abovecaptionskip}{-5pt}
\setlength{\belowcaptionskip}{-10pt}
  \includegraphics[width=\linewidth]{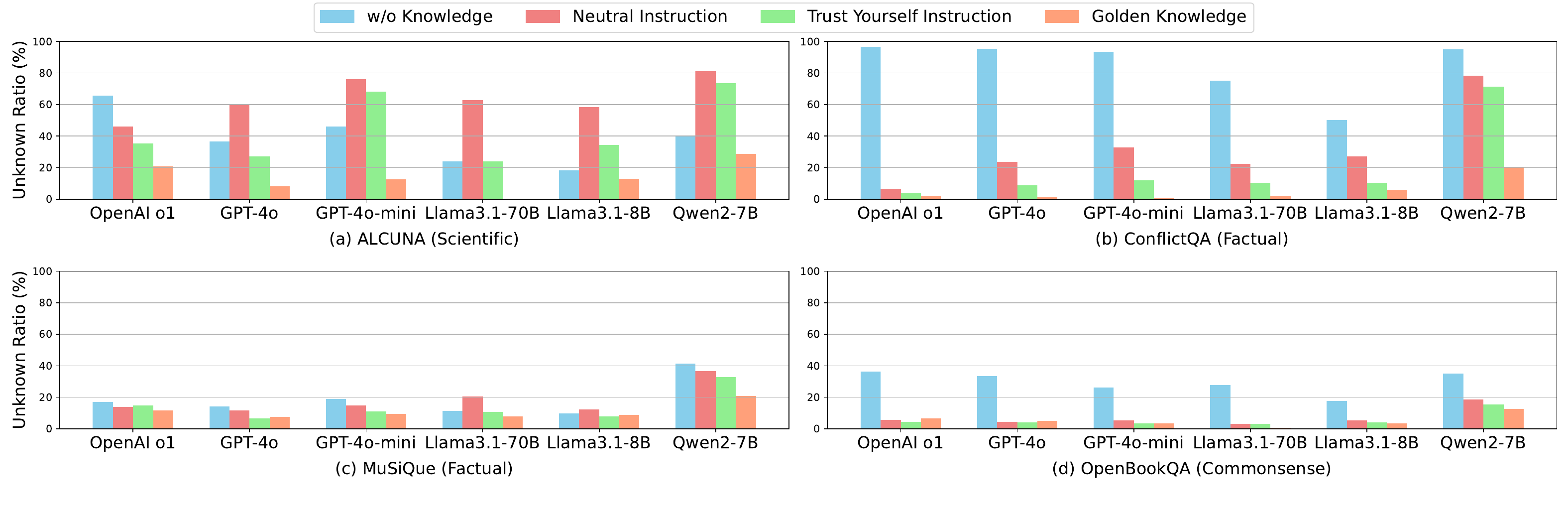}
  \caption {Unknown Ratio for Complementary Reasoning across four datasets. w/o Knowledge and Golden Knowledge means no information and all the required information is given, respectively. The $x$-axis denotes different models.}
  \label{fig:ur_complementary}
\end{figure*}

\begin{figure*}[t]
\setlength{\abovecaptionskip}{-5pt}
\setlength{\belowcaptionskip}{-10pt}
  \includegraphics[width=\linewidth]{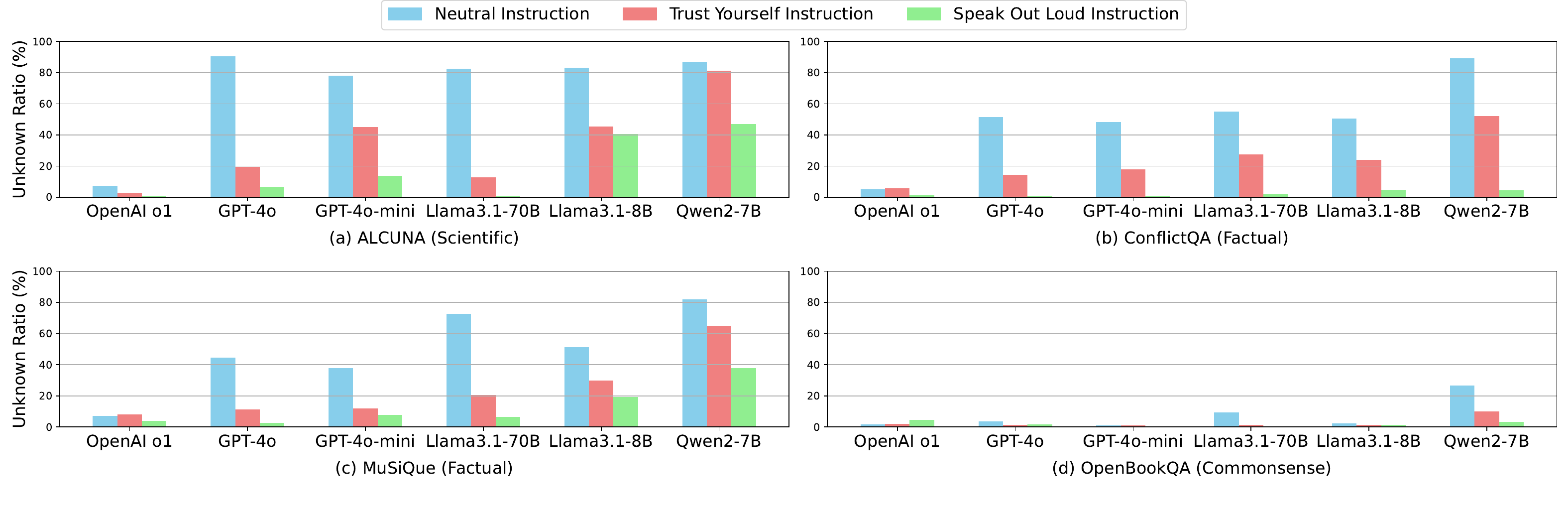}
  \caption {Unknown Ratio for Irrelevant Reasoning.}
  \label{fig:ur_irrelevant}
\end{figure*}

Here we report results by Unknown Ratio $U_R$, which holds similar trend across all datasets and all tested LLMs. Therefore, we do not include all of them in the main body of our paper.
Figure \ref{fig:ur_complementary} illustrates the $U_R$ for Complementary Reasoning. Similarly, Table \ref{tab:tab_ur_conflict1}, \ref{tab:tab_ur_conflict2} and Figure \ref{fig:ur_irrelevant} shows the $U_R$ for Conflicting and Irrelevant Reasoning, respectively.

Echoing discussion in the main body (Section \ref{sec:complementary_reasoning}), the introduction of complementary contextual knowledge may even bring up the uncertainty of models (Figure \ref{fig:ur_complementary} (a)). With our progressively-enforced Reasoning Instructions, LLMs uncertainty go down substantially. But still, although given the sufficiency of knowledge, LLMs tend to choose ``unknown'' to some extent. This again echoes our finding that LLMs cannot fully leverage their knowledge when \ck is present. Also, the $U_R$ for commonsense knowledge is quite low, compared with other types of knowledge, shedding lights on the future improvements.

\begin{table}[!h]
\centering
\begin{minipage}{0.48\linewidth}
    \centering
    \caption{Unknown ratio $U_R$ for Conflicting Reasoning on ALCUNA. NI, TYI, SOLI denotes Neutral, Trust Yourself, Speak Out Loud Instructions, respectively.}
    \vspace{3pt}
    \begin{tabular}{lccc}
        \toprule
        \textbf{Model} & \textbf{NI} & \textbf{TYI} & \textbf{SOLI} \\
        \midrule
        OpenAI o1 & 4.10 & 16.01 & 10.93\\
        GPT-4o & 6.22 & 3.10 & 6.48 \\
        GPT-4o-mini & 2.34 & 1.76 & 1.76 \\
        Llama 3.1-70B & 5.18 & 2.36 & 6.00 \\
        Llama 3.1-8B & 10.99 & 13.51 & 30.58 \\
        Qwen 2-7B & 21.38 & 16.47 & 19.36 \\
        \bottomrule
    \end{tabular}
        \label{tab:tab_ur_conflict1}
\end{minipage}
\hfill
\begin{minipage}{0.48\linewidth}
    \centering
        \caption{Unknown ratio $U_R$ for Conflicting Reasoning on ConflictQA. NI, TYI, SOLI denotes Neutral, Trust Yourself, Speak Out Loud Instructions, respectively.}
\vspace{3pt}
    \begin{tabular}{lccc}
        \toprule
      \textbf{Model} &   \textbf{NI} & \textbf{TYI} & \textbf{SOLI} \\
        \midrule
    OpenAI o1 &    7.29  & 5.13 & 3.49\\
    GPT-4o &    3.33 & 3.93 & 4.39 \\
    GPT-4o-mini &    3.06 & 4.22 & 0.78 \\
    Llama 3.1-70B &    1.98 & 1.99 & 3.02 \\
     Llama 3.1-8B &    3.09 & 5.87 & 9.40 \\
     Qwen 2-7B &    4.56 & 3.36 & 8.35 \\
        \bottomrule
    \end{tabular}
    \label{tab:tab_ur_conflict2}
\end{minipage}
\end{table}

For Conflicting Reasoning as shown in Table \ref{tab:tab_ur_conflict1}, \ref{tab:tab_ur_conflict2}, stronger models show relative lower uncertainty (lower than 7\%). For weaker models in scientific knowledge, such as Llama 3.1-8B and Qwen 2-7B, while our progressively enforced instruction improves their leverage of \pk and encouraging them to trust themselves, the $U_R$ also goes up slightly. We speculate that while the absolute rate of trusting themselves increases, LLMs cannot fully recall their memory of the scientific knowledge distracted by the presence of \ck, even when asked to output their own knowledge.
 
For Irrelevant Reasoning as shown in Figure \ref{fig:ur_irrelevant}, given the Neutral Instruction, LLMs tend to lean on context for reasoning, showing high uncertainty. When the context is irrelevant, although they have the knowledge to answer the question, they still tend to respond ``Unknown'', with Qwen 2-7B reaching 80\% $U_R$. Concerningly, our progressively enforced instructions lower the uncertainty by a large margin, with ``Speak Out Loud Instruction'' almost dropping $U_R$ to lower than 10\% for most LLMs. Also, LLMs exhibit less uncertainty when faced with commonsense knowledge, echoing previous conclusions that LLMs may recall this type of knowledge better than others.

\subsection{When \ck is Required and Mixed with Irrelevant Information}
\label{sec:ck_is_required}

\begin{wraptable}{r}{0.42\linewidth} 
    \setlength{\abovecaptionskip}{-0.39cm}
    \setlength{\tabcolsep}{4pt}
    \caption{Accuracy for Irrelevant Reasoning, requiring \ck~on ALCUNA. Single and \textit{w/} Irr Fact denotes context with only the required fact and with randomly-sampled 1-4 irrelevant facts, respectively.}
    \centering
    \small
    \vspace{3pt}
    \begin{tabular}{lcc}
            \toprule
            \textbf{Model} & \textbf{Single Fact} & \textbf{\textit{w/} Irr Fact} \\
            \midrule
            GPT-4o-mini & 79.78 & 78.81 \\
            Llama 3.1-70B & 86.81 & 87.75 \\
            Llama 3.1-8B  & 71.13 & 65.56 \\
            \bottomrule
        \end{tabular}
        \label{tab:pk_irrelevant}
        \vspace{-.2cm}
\end{wraptable}

As our paper focuses on investigating LLMs ability to leverage \pk, we do not include when \pk~is irrelevant to the question while the required \ck~is mixed with irrelevant knowledge. However, it is also an inevitable case for RAG-systems and shows LLMs ability of \textit{knowledge discrimination or filtering}. We experiment on ALCUNA with its knowledge understanding subset, making sure the knowledge required for the question is not held by LLMs. Results are shown in Table \ref{tab:pk_irrelevant}. It is demonstrated that for knowledge new to the LLMs, their understanding ability is only decent, with Llama 3-8B reaching 71.13\% Accuracy. However, with irrelevant documents, the result goes even higher for some models, showing that the reasoning ability is not greatly affected by the irrelevant knowledge. Compared to the main results in Figure \ref{fig:irrelevant_reasoning}, such result indicates that \textit{LLMs can filter relevant information effectively in the context, but cannot disentangle their own knowledge if it is required for the question.}

\subsection{Case Study}
\label{sec:error_analysis}

To figure out why LLMs suppress their own knowledge at the presence of contextual knowledge. We conduct case study over 200 random samples generated by Llama 3.1-70B with Neutral Instruction (Section \ref{sec:reasoning_instructions}) on ALCUNA \citep{yin-etal-2023-alcuna} where LLMs were able to answer on their own but failed when complementary knowledge was introduced.
Note that we do ask LLMs to use their own knowledge in the
instruction (Section 3.2). We discover that in 93.5\% of cases, LLMs solely rely on the context and even identified the inefficiency of knowledge for reasoning, as an example shown in Table \ref{tab:case_study_2}. Without contextual knowledge, the LLMs can infer the new entity by its name, \textit{e.g., \texttt{dogcat} might have properties similar to dog and cat}. Then, LLMs can reason by the process of elimination to figure out the correct answer. However, the introduce of \ck is likely to limit LLMs reasoning ability to some extent in scientific knowledge. In other cases, the LLMs try to recall their own knowledge, but is eventually distracted by the options or choose the wrong option by hallucination, as examples shown in Table \ref{tab:case_study_1}.

\begin{table}[t]
    \caption{Complementary Reasoning Examples in \textsc{EchoQA}}
    \vspace{1pt}
    \centering
    \begin{tabular}{p{13cm}}
        \toprule      
        \textbf{ALCUNA}
        
        Question: What type of flowers are visited by an organism that is preyed upon by the co-roosting partners of 'Myotis lucifralis'?
        
        0. Haustrum~~1. Picea sitchensis~~2. Scalariogyra~~3. Stellaria media

        Contextual Knowledge: Myotis lucifralis co-roost with Myotis nattereri ......
        
        Ground Truth: 1\\ \midrule
        
       \textbf{ConflictQA}

       Question: What is the person who frequently collaborates with PersonA's occupation?

        0. cartoonist, illustrator~~1. record producer~~2.  film director~~3. unknown

        Contextual Knowledge: Eleanor Davis is the person who frequently collaborates with PersonA.

        Ground Truth: 0 \\ \midrule

       \textbf{MuSiQue}
       
        Question: What administrative territorial entity is the owner of Ciudad Deportiva located?

        0. Tamaulipas~~1. Nuevo Laredo~~2. unknown

        Contextual Knowledge: The Ciudad Deportiva (``Sports City'') is a sports complex in Nuevo Laredo, Mexico....

        Ground Truth: 0 \\ \midrule
        
       \textbf{OpenbookQA}

      Question: Which activity can help a person save money on lunch expenses if they usually eat lunch out?

       Contextual Knowledge: DailyBite offers lunch coupons to frequent customers. FoodieClub is an expensive membership-based dining group...

       Ground Truth: 0  \\ \bottomrule
    \end{tabular}
    \label{tab:complementary_dataset_exp}
\end{table}
        
\begin{table}[t]
    \caption{Conflicting Reasoning Examples in \textsc{EchoQA}. $A_{ck}$ and $A_{pk}$ denotes answer sticking to contextual and parametric knowledge, respectively.}
    \vspace{1pt}
    \centering
    \begin{tabular}{p{13cm}} 
    \toprule
       \textbf{ALCUNA}

        Question: What's the longitude of the location where Phcytodidae is commonly found?

        0. 1.61505 degrees ~~ 1. 93.05 degrees  ~~2. unknown

        Contextual Knowledge: Phcytodidae longitude 1.61505 degrees...

        $A_{ck}$: 0, $A_{pk}$: 1\\ \midrule
       \textbf{ConflictQA} 
       
       Question: Who was the screenwriter for The Hunt?
       
       0. Nick Cuse and Damon Lindelof~~1.David Judah Simon~~2.unknown

       Contextual Knowledge: David Judah Simon is a highly acclaimed screenwriter ......

        $A_{ck}$: 1, $A_{pk}$: 0\\ \midrule
       \textbf{MuSiQue}

       Question: Was \"Green\" recorded as a standalone album after the US tour in late 1977?
       0. Yes~~1. No~~2. unknown

        Contextual Knowledge: \"Green\" was not recorded alone, but rather in conjunction with \"Motivation Radio\"...

                $A_{ck}$: 1, $A_{pk}$: 0\\ \midrule
       \textbf{OpenbookQA}

       Question: In what type of environments is fog more likely to form?
       
       0. Environments with higher humidity, such as marshes.~~1. Environments with lower humidity, such as deserts.~~2. unknown

       Contextual Knowledge: Fog is more likely to form in dry environments with low humidity levels.

       $A_{ck}$: 1 $A_{pk}$: 0\\ \bottomrule
    \end{tabular}
    \label{tab:conflict_dataset_examples}
\end{table}

\begin{table}[t]
    \caption{Irrelevant Reasoning Examples in \textsc{EchoQA}}
    \vspace{1pt}
    \centering
    \begin{tabular}{p{13cm}}
       
        \toprule
         \textbf{ALCUNA}
        
       Question: Does the longitude of Phcytodidae equal 1.61505 degrees?

       0. yes~~1. no~~ 2. unknown

    Contextual Knowledge: Argasidae longitude 88.54616346104861 degrees... Lichida latitude 71.4882053945494 degrees ...
   
    Ground Truth: 1\\ \midrule
\textbf{ConflictQA}

       Question: Who is the author of The Program?

       0.Suzanne Young.~~ 1. Gregg Hurwitz~~2. unknown

       Contextual Knowledge: Anna Todd is a bestselling author who gained popularity...Cormac McCarthy is a highly acclaimed American novelist and playwright...

        Ground Truth: 0\\ \midrule
       \textbf{MuSiQue}

    Question: Where was the Philae lander operated from?
    
    0. DLR's Lander Control Center in Cologne, Germany~~1. A secret facility in the United States~~2. unknown

    Contextual Knowledge: Empire Sports Network was an American regional sports network...Nova Sports is a premium Sports service in Greece and is owned by Forthnet Group...

    0. Eating lunch at DailyBite and using their coupons~~1. Having lunch with FoodieClub~~2. unknown
    
    Ground Truth: 0\\ \midrule
       \textbf{OpenbookQA}
       
Question: What is the primary source of sustenance for predators in an ecosystem?
0. Herbivores or other animals\~~1. Plants and fung~~2. unknown

Contextual Knowledge: Tree roots can push through obstacles like sidewalks as they grow...

Ground Truth: 0\\
    \bottomrule
    \end{tabular}
    \label{tab:irrelevant_dataset_exp}
\end{table}

\begin{table}[t]
    \caption{Examples where complementary knowledge drops the performance on ALCUNA. The LLMs reply excessively on context in these examples.}
    \vspace{1pt}
    \centering
    \begin{tabular} {p{13cm}}
            \toprule
\textbf{Question:} What is the prey of the organism that shares a roosting place with the species 'Myotis lucifralis'?

0. Chara andina~~1. Aldabrachelys~~2. Noctuidae~~3. Geomyidae

\textbf{Ground Truth}: 2

\textbf{Output w/o Contextual Knowledge}:

The species 'Myotis lucifugus' (little brown bat) shares a roosting place with 'Myotis lucifugus', so the correct species to consider is 'Myotis lucifugus'. Now, we need to determine the prey of 'Myotis lucifugus'. Bats are insectivores, and their diet consists of various insects. Among the given options, 'Noctuidae' is a family of moths, which are known to be preyed upon by bats. 

Final Choice: 2

\textbf{Output w/ Contextual Knowledge:}
The organism that shares a roosting place with the species 'Myotis lucifralis' is 'Myotis nattereri'. Since the provided information about 'Myotis lucifralis' does not include the prey of 'Myotis nattereri', but we know 'Myotis lucifralis' preys on 'Noctuidae', among others, and it is likely that 'Myotis nattereri' has a similar diet, but without the exact information about 'Myotis nattereri', the best answer is 'Unknown' since we can't be certain what 'Myotis nattereri' preys on. However, we can make an educated guess that it would be similar to 'Myotis lucifralis', but the options given do not include any prey of 'Myotis lucifralis' other than 'Noctuidae' which is a possible answer if we were certain they had the same prey. In this case, the best answer is 'Unknown' or 'Noctuidae' if we are guessing.

Final Choice: 4\\ \midrule

\textbf{Question: }What is the species that Myotis lucifralis co-roosts with through three intermediate species?

0. Annona pruinosa~~1. Plecotus auritus~~2. Tuxedo drakei~~3. Struvea thoracica

\textbf{Ground Truth}: 1

\textbf{Output w/o Contextual Knowledge}:

 Myotis lucifugus, also known as the little brown bat, is a species of bat known to co-roost with other species. However, the question is specifically asking about co-roosting through three intermediate species. Given the options, I must consider which species is biologically plausible to co-roost with Myotis lucifugus through intermediate species. 
Plecotus auritus, also known as the brown long-eared bat, is known to co-roost with other bat species. Based on the information available, I consider it the most likely candidate among the options provided.

 Final Choice: 1.

\textbf{Output w/ Contextual Knowledge:}

The given information about Myotis lucifralis only mentions that it co-roosts with Myotis nattereri. However, the question asks for a species that Myotis lucifralis co-roosts with through three intermediate species. Since there is no information provided about the co-roosting relationships of Myotis nattereri or any other potential intermediate species, we cannot determine the correct answer based on the given information. Therefore, the correct choice is "Unknown".

Final Choice: 4\\ \bottomrule
    \end{tabular}
    \label{tab:case_study_2}
\end{table}

\begin{table}[t]
    \caption{Examples where complementary knowledge drops the performance on ALCUNA. The LLMs inhibits deep thinking with contextual knowledge compared to ones without knowledge in these examples.}
    \vspace{1pt}
    \centering
    \begin{tabular} {p{13cm}}
        \toprule
     \textbf{Question}: What is eaten by the organism that is parasitized by 'Argasidae', and is itself a predator of another species?
0. Dasypus novemcinctus~1. Ochropleura candelisequa~2. Danaus plexippus~3. Anthurium fragrans~4. Unknown

\textbf{Contextual Knowledge:} 
Argasidae parasitize Homo sapienss......

\textbf{Ground Truth: } 0

\textbf{Output w/o Contextual Knowledge:}
Argasidae is a family of ticks, commonly known as soft ticks. These ticks are ectoparasites that feed on the blood of mammals and birds. Given the options, we need to find an organism that is a predator of another species and can be parasitized by Argasidae. Dasypus novemcinctus, also known as the nine-banded armadillo, is a predator that feeds on various invertebrates and small vertebrates. It can be parasitized by ticks, including Argasidae. Ochropleura candelisequa is a moth, Danaus plexippus is a butterfly, and Anthurium fragrans is a plant, none of which are likely to be parasitized by Argasidae in the context of being a predator of another species.
Final Choice: 0

\textbf{Output w/ Contextual knowledge:}
The question states that the organism is parasitized by 'Argasidae' and is itself a predator of another species. From the given information, we know that 'Argasidae' parasitizes 'Procyon lotor'. We also know that 'Procyon lotor' is eaten by 'Argasidae'. However, the question asks for what is eaten by the organism that is parasitized by 'Argasidae', and is itself a predator of another species, which in this case is 'Procyon lotor'. Since we don't have information about what 'Procyon lotor' eats in the provided data, we can't give a specific answer from the options. However, 'Procyon lotor' is known to eat various small animals including insects, among them, some species of moths such as 'Ochropleura candelisequa' cousins, which is close to option 1.
Final Choice: 1 \\ \hline

 \textbf{Question:}
Uredo has a host, which in turn serves as a host for another organism. What is the host of this second organism? 0. Anthaxia~1. Pelophylax ridibundus
~2. Lolium perenne
~3. Tradescantia~4.Unknown

\textbf{Contextual Knowledge: }Uredo have host Avena sativa......

\textbf{Ground Truth:} 2 

\textbf{Output w/o Contextual Knowledge:}
To answer this question, we need to break it down step by step. Uredo is a genus of rust fungi, which are known to be parasitic on plants. Given that Uredo has a host, which in turn serves as a host for another organism, we can infer that the host of Uredo is likely a plant. Considering the options provided, we should look for the one that is most likely to be a host for Uredo and then a host for another organism.

Looking at the options:
- Anthaxia is a genus of jewel beetles, which are not typically hosts for rust fungi.
- Pelophylax ridibundus is a species of frog, which is not a host for Uredo.
- Lolium perenne is a species of ryegrass, a type of plant.
- Tradescantia is a genus of plants commonly known as spiderworts.

Given that rust fungi typically infect plants, we can narrow down our options to Lolium perenne and Tradescantia. However, we need to consider which of these plants is more likely to be a host for another organism that also serves as a host for Uredo.

Considering the common hosts for Uredo and the possibility of another organism being hosted by the same plant, we should consider Lolium perenne. It is known that some rust fungi infect ryegrass, and this plant can also host other types of organisms.

Final Choice: 2

\textbf{Output w/ Contextual Knowledge:} The hosts of Uredo are listed in the provided information, which includes various plant species. Among the options given, we should look for a plant species that is known to be a host for another organism. Tradescantia is a plant genus that is known to be a host for various organisms, including fungi, insects, and other invertebrates. On the other hand, the other options do not directly relate to known hosts of Uredo or other organisms.

Final Choice: 3\\ \bottomrule
    \end{tabular}
    \label{tab:case_study_1}
\end{table}

\section{Prompt Design}
\label{sec:prompt_design}

We provide a comprehensive list of all the prompts that have been utilized in this study, offering a clear reference for the understanding our experimental approach.

Specifically, in Table \ref{tab:prompt_examples}, \ref{tab:openbook_knowledge} we provide prompt examples adopted to construct \textsc{EchoQA}. In Table \ref{tab:detailed_prompt_reasoning_instructions}, we provide prompt examples for our designed Reasoning Instructions.

\begin{table}[t]
    \caption{Prompt Examples to construct conflicting knowledge~and questions for \textsc{EchoQA}.}
    \vspace{1pt}
    \centering
    \begin{tabular}{>{\raggedright}m{3.2cm}|>{\raggedright\arraybackslash}m{10cm}}
        \toprule
        \textbf{Operation} & \textbf{Prompts} \\ \midrule
        Conflict Knowledge and Question Generation & 
        You are tasked with generating conflicting knowledge based on the given knowledge and creating a related multiple-choice question.
Follow these steps:

1. Generate Conflict Knowledge:
Create a conflicting statement or fact that contradicts the given knowledge. This conflict should challenge the established understanding. \newline
2. Formulate a Multiple Choice Question:
Create a question that asks for the truth regarding the conflicting knowledge. The question should be about the conflict knowledge. Provide two answer choices.\newline Include: \newline
0. One choice that aligns with the given knowledge.\newline
1. One choice that supports the conflicting knowledge.\newline

Your generated conflicting knowledge should contain information with similar volume as the given knowledge. \newline
The Label of choice should be 0. and 1.\newline
Do not output anything after the Desired Output.\newline
Here is the knowledge you should refer:\newline
[Taxon] \newline
Desired Output Format:\newline
Thought: [thought]\newline
Conflicting Knowledge: [new\_knowledge]\newline
Question: [new\_question\_with\_two\_choice]\newline
Let's think step by step.
\\
\bottomrule
    \end{tabular}
    \label{tab:prompt_examples}
\end{table}

\begin{table}[t]
    \caption{Prompt examples for knowledge and question construction for OpenBookQA.}
    \vspace{1pt}
    \centering
    \begin{tabular}{>{\raggedright}m{3.2cm}|>{\raggedright\arraybackslash}m{10cm}}
        \toprule
        \textbf{Operation} & \textbf{Prompts} \\ \midrule
        Commonsense Knowledge Elicitation & 
You are an expert with factual commonsense knowledge on a wide range of topics.
You will be given a multiple-choice question requiring commonsense knowledge to answer.
Your task is to:

1. Output the commonsense knowledge needed to answer the question, making it as concise as possible.

2. Output the answer based on the knowledge.
\newline

For the question and fact provided, provide the needed commonsense knowledge and the answer from the choices.
Desired output format: Knowledge: ["knowledge1", "knowledge2", ...]

Final Answer Choice: [Choice]
DO NOT output anything after the Desired Output.
\newline

Question: Which of these would let the most heat travel through?

Choices: 

A. a new pair of jeans. 

B. a steel spoon in a cafeteria.

C. a cotton candy at a store.

D. a calvin klein cotton hat.
\newline

Knowledge: ["Metal is a thermal conductor.", "Steel is made of metal.","Heat travels through a thermal conductor."]
\newline

Final Answer Choice: B
\newline

Question: [question$\_$stem]

Choices: [formatted$\_$choices]

Knowledge: 
\\ \midrule
Complement Knowledge and Question Generation &
You are an expert with factual commonsense knowledge on a wide range of topics. \newline

Given a multiple-choice question, answer, and related commonsense knowledge, you must create a new question that requires both the given knowledge and new entities/knowledge to answer. \newline

    Step 1: Generate New Entities \newline Create new entities representing concepts or objects related to each choice. These entities must be new and not directly mentioned in the original context. \newline
    Step 2: Create New Knowledge  Create one-hop facts connecting the fake entities to the original choice. Answering the question must require both new and original knowledge. The new knowledge should be a list of strings. \newline
    
    Desired Output \newline {Thought: [thought]}\newline {New Question: [new\_question]}\newline {New Choices: [new\_choices]}\newline {New Knowledge: [new\_knowledge]}\newline {New Answer Key: [new\_answer\_key]} \newline
    Let's think step by step.   \\ \bottomrule
    \end{tabular}
    \label{tab:openbook_knowledge}
\end{table}

\begin{table}[t]
    \caption{Prompt Examples for our designed various levels of Reasoning Instructions.}
    \vspace{1pt}
    \centering
    \begin{tabular}{>{\raggedright}m{3.2cm}|>{\raggedright\arraybackslash}m{10cm}}
        \toprule
w/o Knowledge & You are a powerful question-answering system.\newline
Your task is to use your own commonsense knowledge to answer a multiple-choice question.
If you are unable to answer the question, respond with the ``Unknown'' choice key.\newline
Answer the following question with your choice (A, B, C, D, etc.): [Question]\newline
Desired format: Thought process: [Thought process],  Final Choice: [Final Choice].\newline
Let’s think step by step. \\ \midrule
Neutral / Golden Knowledge Instruction & You are a powerful question-answering system.\newline 
Users will provide some information along with a question. Your task is to combine the information provided by the user with your knowledge to answer the question. 
If you are unable to answer the question, respond with the ``Unknown'' choice key.\newline 
Here is the basic information you can refer:[Taxon]\newline 
Answer the following question with your choice (A, B, C, etc.): [Question]\newline 
Desired format: Thought process: [Thought process], Final Choice: [Final Choice].\newline 
Let’s think step by step. \\ \midrule
Trust Yourself Instruction & You are a powerful question-answering system.\newline 
Users will provide some information along with a question. Your task is to combine the information provided by the user with your knowledge to answer the question.\newline 
If you are unable to answer the question, respond with the ``Unknown'' choice key.\newline 
Here is the basic information you can refer: 
[Taxon]\newline 
Answer the following question with your choice (A, B, C, etc.): [Question]\newline 
Note that the information above is not sufficient to answer the question, you should use your own commonsense knowledge to further reason the final answer.\newline 
Desired format: Thought process: [Thought process], \newline Final Choice: [Final Choice].\newline 
Let’s think step by step. \\ \midrule
Speak Out Loud Instruction &     You are a powerful question-answering system.\newline    Users will provide some information along with a question. Your task is to use the information provided by the user and your own knowledge to answer the question.
    If you are unable to answer the question, respond with the ``unknown'' choice.\newline
    Answer the following question with your choice (A, B, C, etc.): [Question]\newline
    You MUST first output your own knowledge about the question. I am sure you have the knowledge!\newline
    Here is the Given Knowledge:    [Taxon]\newline
    Determine whether the given information about the question conflicts with your own knowledge.\newline
    If you think the Given Knowledge conflicts with your own knowledge, you MUST USE your own knowledge to answer the question.\newline
    Desired format: Thought process: [Thought process], My Own Knowledge: [My Own Knowledge], Final Choice: [Final Choice].\newline
    Let’s think step by step.\\
    \bottomrule
    \end{tabular}
    \label{tab:detailed_prompt_reasoning_instructions}
\end{table}

\end{document}